\documentclass[lettersize,journal]{IEEEtran}

\usepackage{amsmath}
\usepackage{amssymb}
\usepackage{amsfonts}

\usepackage{amsthm}
\usepackage[ruled,vlined]{algorithm2e}
\usepackage{mathtools}
\usepackage{tabularx}
\usepackage{graphicx}
\usepackage{subcaption} 
\usepackage{enumerate}
\usepackage{booktabs}
\usepackage{float}
\usepackage{url}
\usepackage{verbatim}
\usepackage{booktabs} 
\usepackage{multirow}
\usepackage{color}
\usepackage{pifont}
\usepackage[linkcolor=black,citecolor=black,urlcolor=black,colorlinks=true]{hyperref}
\usepackage{cite}
\usepackage{mathrsfs}
\usepackage{xcolor}
\usepackage[normalem]{ulem}
\useunder{\uline}{\ul}{}

\graphicspath{{figures/}}
\IEEEoverridecommandlockouts

\newtheorem{lemma}{Lemma}
\newtheorem{theorem}{Theorem}

\usepackage[normalem]{ulem}

\newcommand{\reduline}{\bgroup\color{black}\markoverwith{\textcolor{red}{\rule[0.3ex]{2pt}{1.5pt}}}\ULon}

\let\oldtwocolumn\twocolumn
\renewcommand\twocolumn[1][]{%
    \oldtwocolumn[{#1}{
    \centering
    \includegraphics[width=0.99\textwidth]{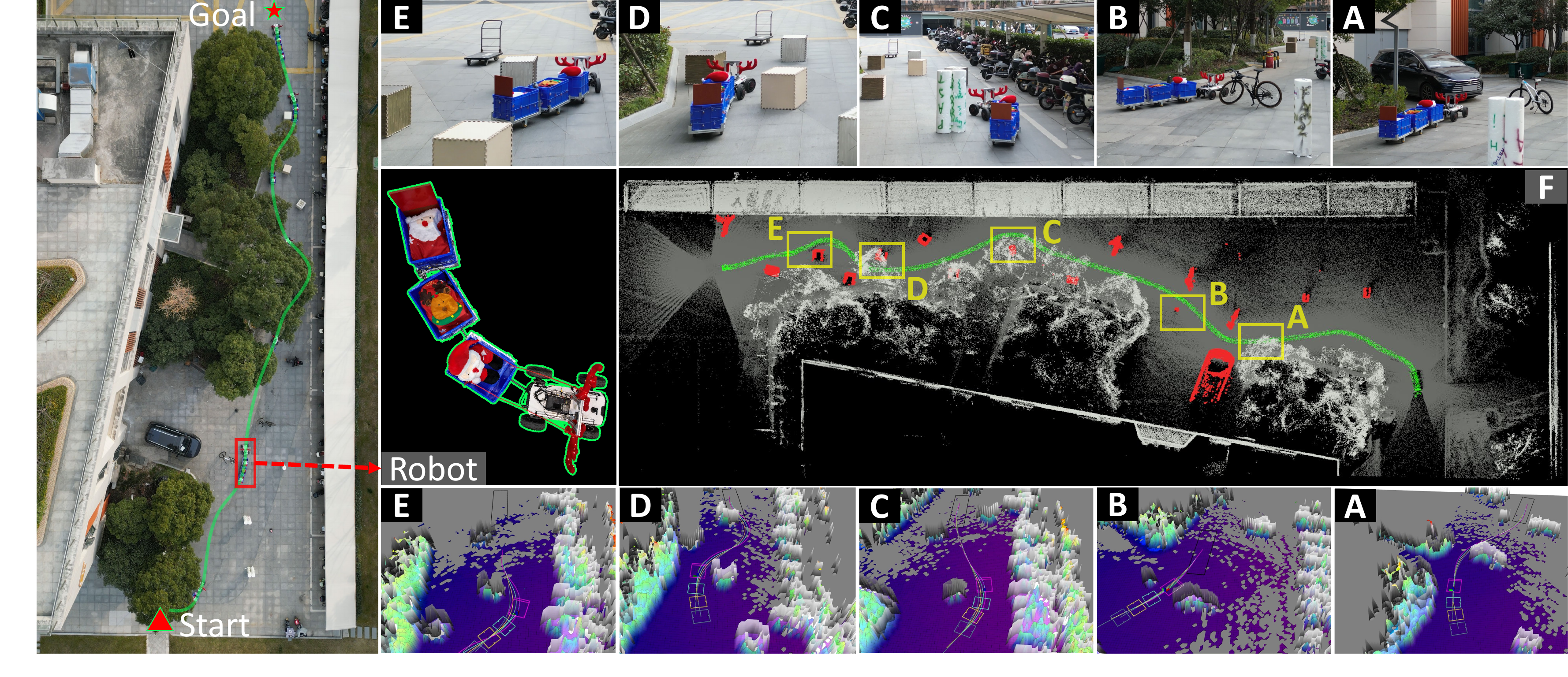}
    \captionof{figure} {A tractor-trailer robot navigating autonomously through narrow lanes. Subfigures A$\sim$E at the top and bottom of the figure correspond to the third-person viewpoint and view from Rviz\cite{rviz} at the moment of the yellow box in the Subfigure F, respectively.}
    \label{fig:head_image}
    \vspace{0.6cm}
    }]
}

\title{Tracailer: An Efficient Trajectory Planner for Tractor-Trailer Robots in Unstructured Environments}

\author{
Long Xu$^{1,2}$, Kaixin Chai$^{2}$, Boyuan An$^{1}$, Shuhang Ji$^{1}$, Zhenyu Hou$^{2}$, Jiaxiang Gan$^{2}$, Qianhao Wang$^{1,2}$, \\Yuan Zhou$^{1,2}$, Xiaoying Li$^{2}$, Junxiao Lin$^{1,2}$, Zhichao Han$^{1,2}$, Chao Xu$^{1,2}$, Yanjun Cao$^{1,2}$, and Fei Gao$^{1,2}$
\thanks{This work was supported by the National Key R\&D Program of China under Grant No. 2023YFB4706600, the Zhejiang Provincial Science and Technology Plan Project under Grant No. 2024C01170 and the National Natural Science Foundation of China under Grant No. 62322314.}
\thanks{$^{1}$State Key Laboratory of Industrial Control Technology, Zhejiang University, Hangzhou 310027, China. \textit{Corresponding author: Fei Gao}}
\thanks{$^{2}$Huzhou Institute of Zhejiang University, Huzhou 313000, China.}
\thanks{E-mail: {\tt\small \{gaolon, fgaoaa\}@zju.edu.cn}}
}

\begin{document}

    \maketitle
    \pagestyle{empty}
    \thispagestyle{empty}

\begin{abstract}
The tractor-trailer robot consists of a drivable tractor and one or more non-drivable trailers connected via hitches. Compared to typical car-like robots, the addition of trailers provides greater transportation capability. However, this also complicates motion planning due to the robot's complex kinematics, high-dimensional state space, and deformable structure. To efficiently plan safe, time-optimal trajectories that adhere to the kinematic constraints of the robot and address the challenges posed by its unique features, this paper introduces a lightweight, compact, and high-order smooth trajectory representation for tractor-trailer robots. Based on it, we design an efficiently solvable spatial-temporal trajectory optimization problem. To deal with deformable structures, which leads to difficulties in collision avoidance, we fully leverage the collision-free regions of the environment, directly applying deformations to trajectories in continuous space. This approach not requires constructing safe regions from the environment using convex approximations through collision-free seed points before each optimization, avoiding the loss of the solution space, thus reducing the dependency of the optimization on initial values. Moreover, a multi-terminal fast path search algorithm is proposed to generate the initial values for optimization. Extensive simulation experiments demonstrate that our approach achieves several-fold improvements in efficiency compared to existing algorithms, while also ensuring lower curvature and trajectory duration. Real-world experiments involving the transportation, loading and unloading of goods in both indoor and outdoor scenarios further validate the effectiveness of our method. The source code is accessible at \href{https://github.com/Tracailer/Tracailer}{\textcolor{magenta}{https://github.com/Tracailer/Tracailer}}.
\end{abstract}

\par
\fontsize{10}{12}\selectfont 
\textbf{\emph{Note to Practitioners}—This paper addresses the challenges of motion planning for tractor-trailer robots, which are crucial in industries like logistics and agriculture. Our approach introduces a lightweight trajectory planning method that efficiently generates safe and time-optimal paths while considering the unique kinematic constraints of these vehicles. By utilizing the environment's collision-free areas, we enhance safety and reduce reliance on initial conditions. While our method shows significant efficiency improvements in simulations and real-world tests, its effectiveness may vary in extremely dynamic environments. Practitioners can easily adopt our open-source code to integrate this planning technique into their operations, improving the efficiency and safety of tractor-trailer systems and paving the way for future advancements in autonomous transportation.}


\section{Introduction}
\label{sec:Introduction}
In recent years, autonomous driving has gained a lot of interest and great growth due to its potential social benefits. When transporting large cargo, attention often shifts to tractor-trailer systems, such as semi-trucks, due to their capacity to carry more through the use of trailers. Tractor-trailer robot is a class of vehicles consisting of a drivable tractor and many unpowered trailers. The trailers are often connected to the tractor or other trailers by hinges, so that the motions of the trailers can be controlled by controlling that of the tractor\cite{dfgen}. 

Motion planning is a crucial component of autonomous navigation frameworks, and tractor-trailer robots are no exception. Similarly to car-like robots, the trajectory planner for tractor-trailer robots aims to generate trajectories that minimize time consumption while satisfying kinematic constraints, enabling smooth and safe navigation to the target area. Although motion planning for car-like robots has been studied relatively extensively, these techniques cannot be directly applied to tractor-trailer robots, since it is necessary to consider not only the state of the tractor but also the state of the trailer when moving. Specifically, compared to car-like robots, the challenges in motion planning for tractor-trailer robots are mainly caused by the following three factors:

\begin{itemize}
    \item[1)]\textbf{Complex kinematics:} The hinge structure and the non-drivability of the trailers introduce more complex kinematic constraints. The coupling of the underactuated constraints of the trailers and the nonholonomic constraints of the tractor\cite{libai_pcoc} requires the planner capable of handling more equation constraints. Meanwhile, it must avoid the jackknife, which will make the entire robot system uncontrollable\cite{jackknife}.
    \item[2)]\textbf{High-dimensional state space:} Each trailer attached to the robot introduces three additional dimensions to the robot in space, i.e. position and yaw angle, which can significantly increase the time consumed by search- or sampling-based path planning algorithms\cite{geo_lat,geo_rrt}. Similarly, for optimization-based methods\cite{libai_pcoc,libai_corridor,ocp_opt,lat_opt,park_opt1}, the increase in dimension of the state leads to an increase in the dimension of the optimization variables, which may result in more memory consumption and slower convergence.
    \item[3)]\textbf{Deformable structure:} The deformability of the tractor-trailer robot, caused by its hinged structure, increases the complexity of collision avoidance. Since the shape of robot changes over time, obstacle avoidance constraints must be applied to each individual unit of the robot. This challenge becomes even more pronounced with multiple trailers, where additional vehicle-to-vehicle collision avoidance constraints must also be considered.
\end{itemize}

To cope with the first two problems, we draw inspiration from the efficiency of polynomial trajectories optimization and differential flatness\cite{mellinger2011minimum} in motion planning for car-like robots\cite{hzctits,hzcnmi}, proposing a new trajectory representation for tractor-trailer robots. We model the spatial-temporal trajectory optimization problem of tractor-trailer robots in continuous space, thus eliminating the discretization of the motion process and substantially improving the efficiency of the optimization. Benefiting from differential flatness\cite{mellinger2011minimum}, we can explicitly derive higher-order states of the robot and reduce the dimensionality of the optimization variables. However, the singularity of differential flatness brought about by kinematics of the tractor may lead to a decrease in the optimization success rate and smoothness of the trajectory, which may make trajectory tracking difficult. Thus, in this work, we introduce \textit{slackened arc length} to eliminate this singularity, improving the success rate of optimization. In addition, this approach makes the feasibility constraints easier to satisfy, leading to a further enhancement in efficiency.

For the third problem of deformed structures, we fully leverage the collision-free regions of the environment, directly applying deformations to trajectories in continuous space. This approach not requires constructing safe regions from the environment using convex approximations through collision-free seed points before each optimization, avoiding the loss of the solution space, thus reducing the dependency of the optimization on initial values, making it easier for the optimizer to find a better solution. With this advantage, to further boost the efficiency, we propose a multi-terminal path search algorithm for tractor-trailer robots, which can searches only in the $SE(2)$ space. Compared to exploration in larger state spaces, our approach achieves nearly an order of magnitude improvement in efficiency with minimal loss in optimality. Contributions of this paper are:
\begin{itemize}
\item [1)]We propose a new trajectory representation for tractor-trailer robots that retains the high efficiency benefits from polynomial representations while eliminating the singularity of differential flatness, further improving the success rate of the trajectory optimization. 
\item [2)]Based on the above trajectory representation, we propose a new optimization-based trajectory planner (Tracailer) for \textbf{trac}tor-tr\textbf{ailer} robots. It directly applies deformations to trajectories in continuous space for ensuring safety, reducing the dependency of optimization on initial values. Meanwhile, an efficient multi-terminal path finder is designed and embedded into it for acquiring the initial values of the problem, achieving higher efficiency without losing much optimality.
\item [3)]To efficiently solve the optimization problem in the planner, we reduce the dimensionality of the optimization variables and constraints based on differential homogeneous mapping and Lagrange multipliers.
\item [4)]Moreover, we build an autonomous navigation system for tractor-trailer robots, including a real-time replanning and mapping module, demonstrating the effectiveness and efficiency of the proposed method in extensive simulations and real-world experiments, such as those shown in Fig.~\ref{fig:head_image}. The software is open-sourced to promote further research in this field.
\end{itemize}

\section{Related Works}
\label{sec:RelatedWork}
Motion planning for tractor-trailer robots can be mainly classified into geometry-based methods and optimization-based methods. 

\subsection{Geometry-based Methods}
Geometry-based methods usually plan only the geometric path and design the path tracking controller to track it. The geometric path is usually generated by search- or sampling-based methods. Search-based methods typically pre-construct a grid map or graph structure that represents the environment, and then perform graph searching within it to find paths~\cite{hybridAs}. For example, Liu et al.~\cite{geo_ga} utilized the conception of equivalent size for obstacles growing and robot shrinking, planning the path by applying a heuristic genetic algorithm. Sun et al.~\cite{geo_prm} built a global compound roadmap by combining the local regular roadmap with the universal probabilistic roadmap, lowering the complexity of the planning computation. To improve the efficiency and path quality, Ljungqvist et al.~\cite{geo_lat} generated a finite set of kinematically feasible motion primitives offline, searching solutions for a general 2-trailer system in a regular state lattice created by these motion primitives. They also proposed a novel parametrization of the reachable state space to make the graph-search problem tractable for real-time applications. Furthermore, by simplifying the design of motion primitives and utilizing reinforcement learning to obtain a well-defined heuristic function, Leu et al. presented an improved A-search guided tree\cite{geometry4} that allows quick off-lattice exploration to find a solution.

Sampling-based methods usually define a space for a path planning task. They maintain a tree structure and iteratively sample in this space, expanding tree nodes according to some criteria, ultimately extracting the trajectories from the tree containing the start and goal. By customizing the sampling space based on motion primitives instead of the space containing all $SE(2)$ states of vehicles in the tractor trailer robot, Cheng et al.~\cite{geometry1}, Lattarulo et al.~\cite{geo_rrt} used the Rapidly-exploring Random Tree\cite{rrt,rrts} (RRT) algorithm to efficiently plan a feasible path that is satisfied with the nonholonomic and mechanical constraints. Moreover, Manav et al.~\cite{geo_crrt} proposed an iterative analytical method, combining it with the Closed-Loop Rapidly Exploring Random Tree (CL-RRT) approach in cascade path planning, allowing the generation of both kinematically feasible and deterministic parking maneuvers with obstacle avoidance.

After obtaining the path by search- or sampling-based methods, the trajectory tracker will choose a piece of path as reference based on the current state of the robot and compute the control command sent to the robot\cite{geometry2}. For example, in work\cite{geometry1}, the path is followed with fuzzy control and Line-of-sight approach\cite{fuzzy}. While in work\cite{geometry3}, a semidefinite programming (SDP) problem is constructed to combine with motion primitive to get the command. Dahlmann et al.~\cite{mpc_opt,mpc_opt2} proposed model predictive controllers to perform local optimizations with Voronoi field\cite{volo} for suboptimal reference trajectories.

\subsection{Optimization-based Methods}
Although geometry-based methods can work in various scenarios, these approach always search in a discrete space, where the optimality of the solution obtained is positively correlated with the duration of the search and the granularity of the space. When information of higher dimensions in the state of the robot is considered to find a more optimal solution, these methods need to explore larger space, whereupon the problem of combinatorial explosion is magnified, severely affecting the efficiency. To find better solutions in such a large space, optimization-based approaches emerged. These methods often model motion planning as an optimization problem and use numerical solvers to solve it, whose initial values often derived from geometric paths obtained by search- or sampling-based methods. Trajectory optimization typically explores optimal solutions in a continuous space, which can take into account information about higher-order states with guarantees of local optimality. Muralidhara et al.~\cite{ocp_opt} modeled the motion planning problem of tractor-trailer robots as Optimal Control Problem (OCP), describing the trajectory by states in discrete timestamps, solving it by direct multiple shooting method. But in obstacle avoidance, this approach only considers not leaving the tracking line of the traffic lane, which cannot be directly applied to unstructured complex environments. Mohamed et al.~\cite{apf_opt} proposed an approach combining artificial potential fields and optimal control theory to achieve a more generalized obstacle avoidance for tractor-trailer robot. However, this method can easily fall into a local optimum when the environment is complex, preventing the robot from reaching the target region. Also using the potential field approach to represent the environment, Wang et al.~\cite{lat_opt2} adopt path integral policy improvement to optimize the paths from geometry based methods, solving the problem of poor performance of graph search methods in narrow scenes the limitations by resolution and search space. Nevertheless, this approach is time-consuming, making it difficult to apply to tasks that require online replanning, such as exploring unknown environments and navigating through dynamically changing scenarios.

The method proposed in this paper is also an optimization-based algorithm. To plan efficiently, we do not adopt discrete states to represent trajectory and model the problem as an OCP, but propose a lightweight, compact and high-order smooth trajectory representation, reducing the dimension of optimization variables and simplifying the problem.

In works\cite{system_opt,system_opt2,lat_opt}, both obstacles and robots are represented by many circles, as this simple form can be easily modeled into OCP. However, real-world environments are complex and cannot be accurately modeled using only circles. 
To represent the environment more concisely and accurately, some works~\cite{libai_nogen, libai_pcoc,apten,park_opt1,basedlibai_opt} use convex polygons to approximate obstacles. They also modeled the motion planning problem for tractor-trailer robots as OCP, considering safety constraints and duration in the optimization. Their methods require applying constraints to each obstacle and each vehicle of the robot at each moment. This feature tightly couples their computational complexity to the environment, limiting the efficiency of planning in obstacle-dense environments. To address this issue, some works\cite{libai_corridor,geometry5} shift to Safe Travel Corridors (STC) as obstacle avoidance constraints. The constraints imposed by this method are only related to the trajectory length and not to the density of obstacles. However, the classical safe corridors must be computed before trajectory optimization, since they are constructed based on the initial values of the optimization. For motion planning of a tractor-trailer robot, this will take a lot of time, especially if the number of trailers is large, since the safety corridors need to be built for each vehicle of the robot separately. The second disadvantage of the safe corridor is its high dependence on initial value, while a good one requires much more time due to the high state dimension and complex kinematics of the robot. Usually, the process of safe corridor generation selects some state points of a trajectory as seeds, which will affect the location and size of the generated safe corridors. Thus in some scenarios where it is more difficult to obtain a good initial value of the trajectory, e.g., the tractor-trailer robot needs to make a turn in a narrow environment, the feasible region constrained by the safe corridors may not be able to contain sufficiently good solutions. To reduce the dependency of the optimization on initial values, Li et al.\cite{libai_astar} proposed a multi-stage method that only uses paths from A* as the initial value for trajectory optimization of tractor-trailer robots traversing a curvy tunnel, improving some of the efficiency for this task.

In this paper, we fully leverage the collision-free regions of the environment, directly applying deformations to trajectories in continuous space, which can avoid the above-mentioned problems brought by safe corridors. In known scenarios, the process for environment can be pre-done to adapt our method, while in unknown environments we can also use separately opened thread for maintenance without occupying the time of trajectory planning. 

\section{Planning Framework}
\label{sec:Planning Framework}

\begin{figure}	
    \centering
    \includegraphics[width=1.0\columnwidth]{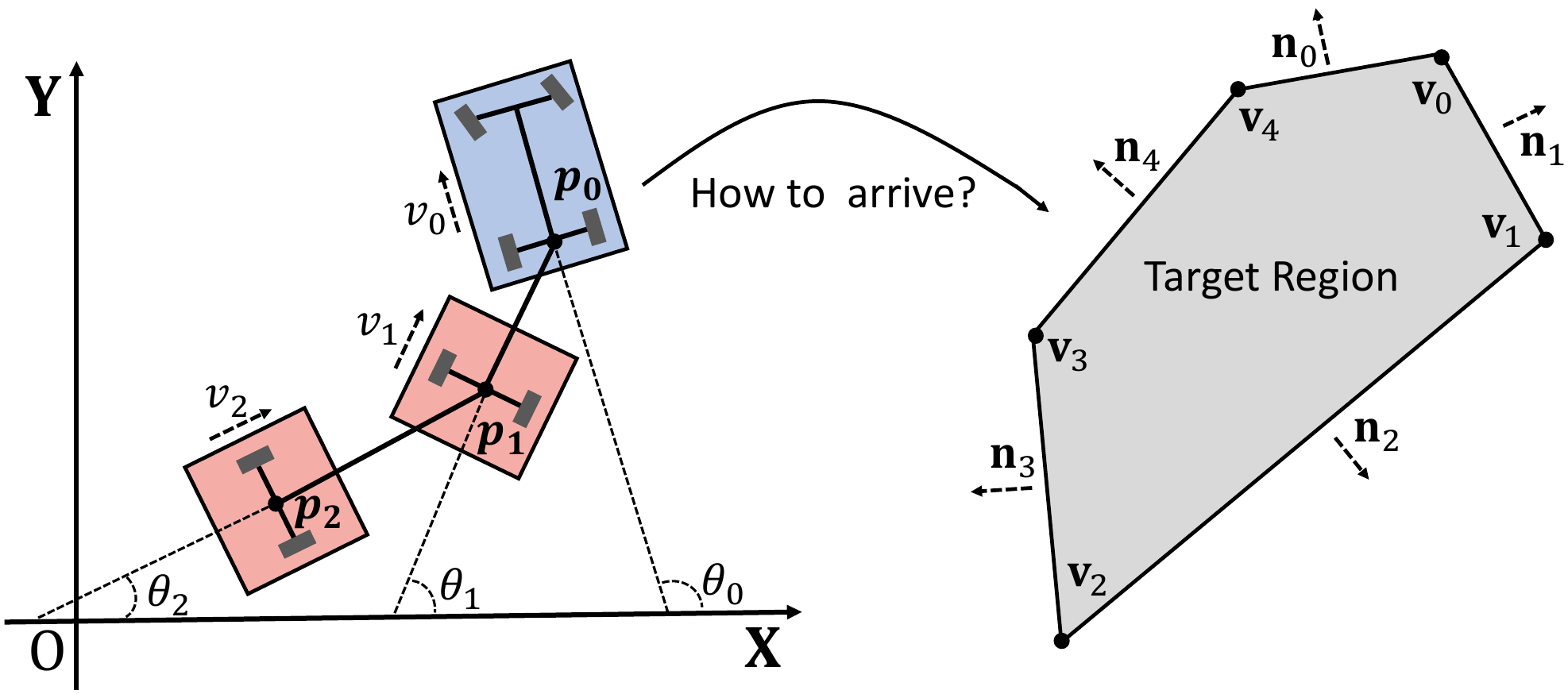}
	\caption{An example of a tractor-trailer robot ($\text N=2$). The first trailer is hooked to the rear center of the tractor, while the others are hooked to the center of the previous one.}\label{fig:kinematics}
\end{figure}
In this work, we mainly focus on tractor-trailer robots whose tractor has an Ackermann chassis, as shown in Fig. \ref{fig:kinematics}. Using the simplified bicycle model\cite{bicycle} and letting the rear center of the tractor of the robot with $\text N$ trailers is $\boldsymbol{p}_0=[x_0,y_0]^\text T$, and the position of each trailer is denoted by $\boldsymbol{p}_i=[x_i,y_i]^\text T,i=1,2,...,\text N$, we can write the kinematic model of the robot as follows:
\begin{align}
\dot x_0(t)&=v_0(t)\cos\theta_0(t),\label{eq:kin_begin}\\
\dot y_0(t)&=v_0(t)\sin\theta_0(t),\label{eq:kin_y}\\
\dot v_0(t)&=a(t),
\end{align}
\begin{align}
\dot \theta_0(t)&=v_0(t)\frac{\tan\delta(t)}{L_0},\label{eq:kin_th}\\
\dot \theta_1(t)&=\frac{v_0(t)\sin(\theta_0(t)-\theta_1(t))}{L_1},\\
\vdots\nonumber\\
\dot \theta_i(t)&=\frac{v_{i-1}(t)\sin(\theta_{i-1}(t)-\theta_i(t))}{L_i},\label{eq:kin_cstr}\\
\vdots\nonumber\\
\dot \theta_{\text{N}}(t)&=\frac{v_{\text N-1}(t)\sin(\theta_{\text N-1}(t)-\theta_{\text N}(t))}{L_{\text N}},\label{eq:kin_end}
\end{align}
where $t\in(0,+\infty)$ denotes the time, $v_0,a,\delta,L_0$ are the velocity, longitudinal acceleration, steering angle, and wheelbase length of the tractor, respectively. $\theta_i, i=0,1,...,\text N$ is the yaw angle of each vehicle in the world frame. For convenience, we use $\boldsymbol{\theta}(t)=[\theta_1(t), \theta_2(t),...,\theta_{\text N}(t)]$ for the yaw angles of all trailers. $v_i=v_{i-1}\cos(\theta_{i-1}-\theta_i), L_i, i=1,...\text N$ are the velocity and length of the rigid link of each trailer, respectively.

Our goal is to plan a kinematically feasible and collision-free trajectory with the initial state of the robot $[x_0^\text{init},y_0^\text{init},v_0^\text{init},\theta_0^{\text{init}},\boldsymbol{\theta}^\text{init}]^\text T$ where $\boldsymbol{\theta}^\text{init}=[\theta_1^\text{init},\theta_2^\text{init},...,\theta_\text{N}^\text{init}]$ and information about the obstacles $\textbf{Info}_c$, such that the robot can follow this trajectory and eventually arrive within the assigned target region. In this work, point cloud are used to characterize the obstacles information as it can be directly acquired from LiDAR. Besides, we utilize a 2D convex polygon $\textbf{Info}_e$ to describe the target region, which is a more general condition compared to the work\cite{libai_corridor} and can also brings convenience for trajectory optimization.

\subsection{Trajectory Parameterization}
\label{subsec:Trajectory Parameterization}
Differential flatness and polynomial representations have brought a significant improvement in the efficiency of trajectory generation for car-like robots\cite{hzctits}. The trajectory representation in this work is also inspired by these. For the robot like the one in Fig. \ref{fig:kinematics}, Deligiannis et al.\cite{dftrailer} have demonstrated it is a differential flat system with $\boldsymbol{p}_\text{N}$ as flat output. However, the order of the derivatives involved in the derivation process of physical variables such as the longitudinal acceleration and steering angle of the tractor is positively correlated with the number of trailers. Thus the nonlinearity of the system will sharply strengthen as the number of trailers increases. Meanwhile, this method cannot be extended to more general tractor-trailer robot system, where the trailer $\boldsymbol{p}_{i+1}$ is not hooked directly to the center $\boldsymbol{p}_{i}$ of the previous one ($i=1,2,...,\text N$), but at a distance from this point.\cite{dfgen}

To simplify the robot state representation and to allow the method to be extended to more general tractor-trailer systems, we use $[x_0(t),y_0(t),\boldsymbol{\theta}(t)]^\text T$ to represent the trajectory of the robot. Combining with Eq.(\ref{eq:kin_begin})$\sim$Eq.(\ref{eq:kin_end}), we can compute some useful physical variables as follows:
\begin{align}
\theta_0(t)&=\text{atan2}(\dot y_0(t),\dot x_0(t)),\label{eq:df_begin}\\
v_0(t)&=\sqrt{\dot x_0^2(t)+\dot y_0^2(t)},
\end{align}
\begin{align}
a(t)&=\frac{\dot x_0(t)\ddot x_0(t)+\dot y_0(t)\ddot y_0(t)}{\sqrt{\dot x_0^2(t)+\dot y_0^2(t)}},\\
\kappa(t)&=\frac{\dot x_0(t)\ddot y_0(t)-\dot y_0(t)\ddot x_0(t)}{\sqrt{(\dot x_0^2(t)+\dot y_0^2(t))^3}},\label{eq:df_end}
\end{align}
where $\kappa(t)$ is the curvature of the trajectory. However, when the robot is stationary, i.e., there is a moment $t=t_s$, such that $v_0(t_s) = 0$, the above part of the equations will have a zero denominator, which is known as the singularity of differential flatness\cite{hzcnmi}. To avoid this problem, existing works usually add small positive constants to the denominator\cite{hzctits} or set a minimum velocity constraint\cite{zmkicra}. These can lead to larger trajectory tracking errors, as well as making it difficult for the optimizer to converge to a solution that satisfies the kinematic constraints, especially the curvature constraint. To demonstrate this point, we introduce a model predictive control (MPC)-based controller to measure the tracking error. Details can be found in Sec.\ref{subsec:Cases Study} and Sec.\ref{subsec:Benchmark Comparisons}.

To solve this problem, inspired by the work\cite{arcijrr} that utilizes arc length $s(t)$ to represent trajectories, we introduce \textit{slackened arc length trajectory} $\mathfrak{s}(t)$ and a constraint to avoid singularities. Where $s$ and $\mathfrak{s}$ are both functions of the time $t$. For brevity, In the rest of the paper, we will simplify them to writing $s$ and $\mathfrak{s}$ instead of $s(t)$ and $\mathfrak{s}(t)$ when necessary. According to the work\cite{arcijrr}, the arc length $s(t)$ is satisfied: $s(0)=0,\dot s(t)=\sqrt{\dot x_0^2(t)+\dot y_0^2(t)}=v_0(t)\ge0$. Also, Eq.(\ref{eq:kin_begin}) and Eq.(\ref{eq:kin_y}) are rewritten as:
\begin{align}
\dot x_0(s)\dot s(t)&=\dot s(t)\cos\theta_0(t),\\
\dot y_0(s)\dot s(t)&=\dot s(t)\sin\theta_0(t).
\end{align}
Thus, we can rewrite Eq.(\ref{eq:df_begin})$\sim$Eq.(\ref{eq:df_end}) as follows:
\begin{align}
\theta_0(t)&=\text{atan2}(\dot y_0(s)\dot s(t),\dot x_0(s)\dot s(t)),\\
v_0(t)&=\dot s(t),\\
a(t)&=\ddot s(t),\\
\kappa(t)&=\frac{(\dot x_0(s)\ddot y_0(s)-\dot y_0(s)\ddot x_0(s))\dot s^3(t)}{\dot s^3(t)}.
\end{align}
Due to the continuity of the robot's state, for the moment $t=t_s$ with $\dot s(t_s) = 0$, we only need to define: $\theta_0(t_s)=\lim_{t\rightarrow t_s}\theta_0(t),\kappa(t_s)=\lim_{t\rightarrow t_s}\kappa(t)$ and make sure that $\dot x_0^2(s(t_s))+\dot y_0^2(s(t_s))>0$. At this point, we have solved the singularity problem, but introduced the following constraints:
\begin{align}
s(t)&\ge0,\\
\dot s(t)&\ge0,\\
\dot x_0^2(s)+\dot y_0^2(s)&=1 \quad \text{when }\dot s(t)\neq0,\label{con:dots}\\
\dot x_0^2(s)+\dot y_0^2(s)&>0 \quad \text{when }\dot s(t)=0.
\end{align}
For tractor, it is rare for the forward velocity to be zero. Thus, using polynomials to represent the trajectory poses a greater challenge to the optimizer because of the need to impose equation constraint (\ref{con:dots}) at each timestamp. 

To slacken the constraint (\ref{con:dots}) and facilitate optimization, we define a new variable \textit{slackened arc length} $\mathfrak{s}(t)$ to replace the arc length $s(t)$. In this paper, we follow a part of the definition of the arc length $s(t)$ by setting $\mathfrak{s}(0)=0$ and $\dot{\mathfrak{s}}(t)\ge0$. Similarly, we can rewrite Eq.(\ref{eq:kin_begin}) and Eq.(\ref{eq:kin_y}) as:
\begin{align}
\dot x_\text{new}(\mathfrak{s}(t))\dot{\mathfrak{s}}(t)&=v_0(t)\cos\theta_0(t),\label{eq:22}\\
\dot y_\text{new}(\mathfrak{s}(t))\dot{\mathfrak{s}}(t)&=v_0(t)\sin\theta_0(t),\label{eq:23}
\end{align}
Where both $x_\text{new}$ and $y_\text{new}$ are differentiable functions of $\mathfrak{s}$, $\mathfrak{s}$ is a differentiable function of time $t$, and $x_\text{new}(\mathfrak{s}(t)) = x_0(t),y_\text{new}(\mathfrak{s}(t))=y_0(t)$. Note that Eq.(\ref{eq:22}) and Eq.(\ref{eq:23}) can be easily proved to be equivalent to Eq.(1) and Eq.(2), respectively, by the chain rule. Thus, for brevity, we will still use the symbols $x_0(\mathfrak{s}(t))$ and $y_0(\mathfrak{s}(t))$ instead of the symbols $x_\text{new}(\mathfrak{s}(t))$ and $y_\text{new}(\mathfrak{s}(t))$ in the subsequent parts of the paper.

Subsequently, adding the constraint $\dot x_0^2(\mathfrak{s})+\dot y_0^2(\mathfrak{s})\ge\delta_+$, where $\delta_+>0$ is a constant, and using the technique mentioned in the previous paragraph related to the continuity of the robot's state, we can rewrite Eq.(\ref{eq:df_begin})$\sim$Eq.(\ref{eq:df_end}) as follows:
\begin{align}
\theta_0(t)=&\text{atan2}(\dot y_0(\mathfrak{s}),\dot x_0(\mathfrak{s})),\\
v_0(t)=&\dot{\mathfrak{s}}(t)\sqrt{\dot x_0^2(\mathfrak{s})+\dot y_0^2(\mathfrak{s})},\\
a(t)=&\ddot{\mathfrak{s}}(t)\sqrt{\dot x_0^2(\mathfrak{s})+\dot y_0^2(\mathfrak{s})}\nonumber\\
&+\dot{\mathfrak{s}}^2(t)\frac{\dot x_0(\mathfrak{s})\ddot x_0(\mathfrak{s})+\dot y_0(\mathfrak{s})\ddot y_0(\mathfrak{s})}{\sqrt{\dot x_0^2(\mathfrak{s})+\dot y_0^2(\mathfrak{s})}},\\
\kappa(t)=&\frac{\dot x_0(\mathfrak{s})\ddot y_0(\mathfrak{s})-\dot y_0(\mathfrak{s})\ddot x_0(\mathfrak{s})}{\sqrt{(\dot x_0^2(\mathfrak{s})+\dot y_0^2(\mathfrak{s}))^3}}.
\end{align}

Same as in the works\cite{uneven,hzctits} to consider human comfort by minimizing integration of jerk\cite{minjerk}, based on the optimality conditions proved by Wang et al.\cite{minco} for the multi-stage control effort minimization problem, we choose quintic piecewise polynomials with four times continuously differentiable at the segmented points as the representation for trajectories of the robot. Introducing $\mathfrak{s}(t)$, each piece of trajectory is denoted as:
\begin{align}
x_{j0}(\mathfrak{s})&=\boldsymbol{c}_{x_j}^\text{T}\gamma(\mathfrak{s})\quad &\mathfrak{s}\in[0,\mathfrak{s}_j(T_p)-\mathfrak{s}_j(0)],\\
y_{j0}(\mathfrak{s})&=\boldsymbol{c}_{y_j}^\text{T}\gamma(\mathfrak{s})\quad &\mathfrak{s}\in[0,\mathfrak{s}_j(T_p)-\mathfrak{s}_j(0)],\\
\mathfrak{s}_{j}(t)&=\boldsymbol{c}_{\mathfrak{s}_j}^\text{T}\gamma(t)\quad &t\in[0,T_p],\\
\theta_{ki}(t)&=\boldsymbol{c}_{\theta_{ki}}^\text{T}\gamma(t)\quad &t\in[0,T_\theta],
\end{align}
where $j=1,2,...,\text{M};k=1,2,...,\Omega$ is the index of piecewise polynomial, $T_*,*=\{p,\theta\}$ is the duration of a piece of the trajectory, $c_*\in\mathbb{R}^6,*=\{x_j,y_j,\mathfrak{s}_j,\theta_{ki}\}$ is the coefficient of polynomial, $\gamma(*)=[1,*,*^2,...,*^5]^\text{T},*=\{t,\mathfrak{s}\}$ is the natural base.

\subsection{Optimization Problem}
\label{subsec:Optimization Problem}
In this paper, we formulate the trajectory optimization problem for tractor-trailer robots as:
\begin{align}
\min_{\boldsymbol{C},\boldsymbol e,T_{f}}&f(\boldsymbol{C},\boldsymbol e,T_{f})=\int_0^{\Sigma_{j=1}^\text{M}\mathfrak{s}_j(T_p)}\boldsymbol{j}_p(\mathfrak{s})^\text{T}\boldsymbol W_p\boldsymbol{j}_p(\mathfrak{s})d\mathfrak{s}\\
&\quad\quad\quad\quad\quad\ +\int_0^{T_f}\boldsymbol{j}_\theta(t)^\text{T}\boldsymbol W_\theta\boldsymbol{j}_\theta(t)dt+\rho_tT_f\label{problem:opt}\\
&s.t.\ \ \dot\theta_i(t)L_i=v_{i-1}(t)\sin(\theta_{i-1}(t)-\theta_i(t)),\label{con:eqc}\\
&\quad\quad \boldsymbol{M}_{p}(\boldsymbol{S})\boldsymbol{c}_{p}=\boldsymbol{b}_{p}(\boldsymbol{P},\boldsymbol{e}),\label{con:mincop}\\
&\quad\quad \boldsymbol{M}_{\mathfrak{s}}(T_p)\boldsymbol{c}_{\mathfrak{s}}=\boldsymbol{b}_{\mathfrak{s}}(\boldsymbol{S}),\label{con:mincos}\\
&\quad\quad \boldsymbol{M}_{\theta}(T_\theta)\boldsymbol{c}_{\theta}=\boldsymbol{b}_{\theta}(\boldsymbol{\Theta},\boldsymbol{e}),\label{con:mincot}\\
&\quad\quad T_f>0,\quad\mathfrak{s}(t)\ge0,\quad\dot{\mathfrak{s}}(t)\ge0,\label{con:vsgeq0} \\
&\quad \quad \dot x_0^2(\mathfrak{s})+\dot y_0^2(\mathfrak{s})\ge\delta_+,\label{con:slackv}\\
&\quad \quad \lvert\theta_{i-1}(t)-\theta_{i}(t)\rvert\leq\delta\theta_{\text{max}},\label{con:dyn_begin}\\
&\quad \quad v_0^2(t)\leq v_\text{mlon}^2,\quad a^2(t)\leq a_\text{mlon}^2,\\
&\quad \quad a_\text{lat}^2(t)\leq a_\text{mlat}^2,\quad\kappa^2(t)\leq \kappa_{\text{max}}^2,\label{con:dyn_end}\\
&\quad \quad \boldsymbol{\mathcal{G}}_e(\boldsymbol e,\textbf{Info}_e)\leq\boldsymbol0,\label{con:end}\\
&\quad \quad \boldsymbol{\mathcal{G}}_c(\boldsymbol{C},\boldsymbol e,T_{f},\textbf{Info}_c)\leq\boldsymbol{0},\label{con:safe}
\end{align}
where 
$\boldsymbol{c}_p=[[\boldsymbol{c}_{x_1},\boldsymbol{c}_{y_1}]^\text T,[\boldsymbol{c}_{x_2},\boldsymbol{c}_{y_2}]^\text T,...,[\boldsymbol{c}_{x_\text M},\boldsymbol{c}_{y_\text M}]^\text T]^\text{T}\in\mathbb{R}^{6\text M\times2}$,$\boldsymbol{c}_\mathfrak{s}=[\boldsymbol{c}_{\mathfrak{s}_1}^\text T,\boldsymbol{c}_{\mathfrak{s}_2}^\text T,...,\boldsymbol{c}_{\mathfrak{s}_\text M}^\text T]^\text T\in\mathbb{R}^{6\text M\times1}$,$\boldsymbol{c}_\theta=[\boldsymbol{c}_{\theta_1}^\text T,\boldsymbol{c}_{\theta_2}^\text T,...,\boldsymbol{c}_{\theta_\Omega}^\text T]^\text T\in\mathbb{R}^{6\Omega\times\text N}$, $\boldsymbol{c}_{\theta_k}=[\boldsymbol{c}_{\theta_{11}},\boldsymbol{c}_{\theta_{12}},...,\boldsymbol{c}_{\theta_\text{1N}}]\in\mathbb{R}^{6\times\text N}$ are coefficient matrices. $\boldsymbol{C}=\{\boldsymbol{c}_p,\boldsymbol{c}_\mathfrak{s},\boldsymbol{c}_\theta\}$. For the other optimization variables, $\boldsymbol{e}=[x_0^\text{fina},y_0^\text{fina},\theta_0^{\text{fina}},\boldsymbol{\theta}^\text{fina}]^\text T$, $T_f=\text MT_p=\Omega T_\theta$ denote the end state of the robot and total duration of the trajectory, respectively, where $\boldsymbol{\theta}^\text{fina}=[\theta_1^\text{fina},\theta_2^\text{fina},...,\theta_\text{N}^\text{fina}]$.

In the objective function $f(\boldsymbol{C},\boldsymbol e,T_{f})$, $\boldsymbol{j}_p(\mathfrak{s})=[x_0^{(3)}(\mathfrak{s}),y_0^{(3)}(\mathfrak{s})]^\text T$ and $\boldsymbol{j}_{\mathfrak{s}\theta}(t)=[\mathfrak{s}^{(3)}(t),\boldsymbol{\theta}^{(3)}(t)]^\text T$ denote the jerk of the trajectories. Their weighted square integral represents the smoothness of the trajectory. $\boldsymbol W_p\in\mathbb{R}^{2\times2},\boldsymbol W_{\mathfrak{s}\theta}\in\mathbb{R}^{(\text{N+1})\times(\text{N+1})}$ are diagonal positive matrices representing the weights. $\rho_t>0$ is a constant to give the trajectory some aggressiveness.

Eq.(\ref{con:eqc}) denotes the kinematic constraints (\ref{eq:kin_cstr}). Eq.(\ref{con:mincop})$\sim$Eq.(\ref{con:mincot}) denote the combinations of the continuity constraints mentioned in last subsection and boundary conditions of the trajectory: 
\begin{align}
[x_0(0),\dot x_0(0)]&=[x_0^\text{init},\cos\theta_0^\text{init}],\\
[y_0(0),\dot y_0(0)]&=[y_0^\text{init},\sin\theta_0^\text{init}],\\
[\mathfrak{s}(0),\dot{\mathfrak{s}}(0)]&=[0,v_0^\text{init}],\\
[\theta_i(0),\dot\theta_i(0)]&=[\theta_i^\text{init},\frac{v_{i-1}^\text{init}\sin\theta_{d(i-1)}^\text{init}}{L_i}],\\
[x_0(\mathfrak{s}_f),\dot x_0(\mathfrak{s}_f)]&=[x_0^\text{fina},\cos\theta_0^\text{fina}],\\
[y_0(\mathfrak{s}_f)),\dot y_0(\mathfrak{s}_f)]&=[y_0^\text{fina},\sin\theta_0^\text{fina}],\\
[\mathfrak{s}(T_f),\dot{\mathfrak{s}}(T_f)]&=[\mathfrak{s}_f,0],\\
[\theta_i(T_f),\dot\theta_i(T_f)]&=[\theta_i^\text{fina},0],
\end{align}
where $\theta_{d(i-1)}^\text{init}=\theta_{i-1}^\text{init}-\theta_i^\text{init}$, $v_{i}^\text{init}=v_{i-1}^\text{init}\cos\theta_{d(i-1)}^\text{init}$, $i=1,2,...,\text N$. $\boldsymbol{P}\in\mathbb{R}^{2\times(\text M-1)},\boldsymbol{\Theta}\in\mathbb{R}^{\text N\times(\Omega-1)}$ are the segment points of the tractor and trailers trajectories, respectively. $\mathfrak{s}_f=\lVert\boldsymbol{S}\rVert_1$ and $\boldsymbol{S}\in\mathbb{R}_{>0}^{\text M\times1}$ denote the total length of the \textit{slackened arc length trajectory} and the length of each segment, respectively. Thus $\textbf{M}_{p},\textbf{M}_{\mathfrak{s}}\in\mathbb{R}^{(6\text M-2)\times6\text M}$, $\textbf{M}_{\theta}\in\mathbb{R}^{(6\Omega-2)\times6\Omega}$. In this work, we set $\dot x_0^2(0)+\dot y_0^2(0)=1$ and $\delta_+=0.9$, which is enough for optimization.

Conditions (\ref{con:vsgeq0})$\sim$(\ref{con:slackv}) denote the positive duration and the constraints brought by the introduction of $\mathfrak{s}(t)$ mentioned in the last subsection. Conditions (\ref{con:dyn_begin})$\sim$(\ref{con:dyn_end}) are dynamic feasibility constraints, including avoiding jackknife\cite{jackknife}, limitation of longitudinal velocity $v_0(t)$, longitude acceleration $a(t)$, latitude acceleration $a_\text{lat}(t)=v_0^2(t)\kappa(t)$, and curvature $\kappa(t)$, where $\delta\theta_\text{max},v_\text{mlon},a_\text{mlon},a_\text{mlat},\kappa_{\text{max}}$ are constants. 

Conditions (\ref{con:end}) denote the constraint of the end state of the robot. In this paper, we set the vertices of the 2D convex polygon denoting the target region be $\textbf{v}_\lambda$, and the normal vectors of corresponding edges toward the outside of the polygon be $\textbf{n}_\lambda,\lambda=0,1,..,\text N_e-1$, as the example of convex pentagon given in the right part of Fig. \ref{fig:kinematics}. We only need to ensure that each vertices of each vehicle of the robot are within this region. Let the position of each vertex of each vehicle in the respective body frame be $\boldsymbol{p}_{i\mu},i=0,1,...,\text N,\mu=1,2,3,4$, we can obtain $4\times(\text N+1)\times(\text N_e-1)$ constraints: $\textbf{n}_\lambda^\text{T}(\boldsymbol{R}_i^\text{fina}\boldsymbol{p}_{i\mu}+\boldsymbol{p}_i^\text{fina}-\textbf{v}_\lambda)\le0$, where
\begin{align}
\boldsymbol{p}_0^\text{fina}&=[x_0^\text{fina},y_0^\text{fina}]^\text T,\\
\boldsymbol{p}_i^\text{fina}&=\boldsymbol{p}_{i-1}^\text{fina}-\boldsymbol{R}_i^\text{fina}[L_i,0]^\text T& i=1,2,...,\text N,\\
\boldsymbol{R}_i^\text{fina}&=\left[\begin{matrix}
\cos\theta_i^\text{fina} & -\sin\theta_i^\text{fina}\\
\sin\theta_i^\text{fina} & \cos\theta_i^\text{fina}
\end{matrix}\right]\quad &i=0,1,...,\text N.
\end{align}
In Conditions (\ref{con:end}), $\boldsymbol0=[0,0,...,0]^\text T\in\mathbb{R}^{4(\text N+1)(\text N_e-1)}$ and the inequality is taken element-wise.

Conditions (\ref{con:safe}) denote the safety constraints. In this paper, we use 3D point clouds $\textbf{Info}_c\in\mathbb{R}^{3\times\text N_c}$ to represent the environment, further extracting obstacle information from it, where $\text N_c$ is the number of the points. Subsequently, we construct a Signed Distance Field (SDF) to fuse the information of collision-free region and obstacles. In SDF, the value at each state of the space is the distance to the edge of the nearest obstacle, where the value inside the obstacle is negative, which can be used to directly deform the trajectory of the robot. For example, assume that the vehicles of the robot can be wrapped by circles of radius $r_i$ and centers $\boldsymbol{p}_{ci}(t),i=0,1,...,\text N$, respectively, $\mathscr{S}(\boldsymbol{p}):\mathbb{R}^2\mapsto\mathbb{R}$ denotes the value of the SDF, i.e., the signed distance from position $\boldsymbol{p}$ to the edge of the nearest obstacle. We can then impose following $\text N+1$ constraints on the optimization problem: $\mathscr{S}(\boldsymbol{p}_{ci}(t))>r_i,i=0,1,...,\text N$, where 
\begin{align}
\boldsymbol{p}_0(t)&=[x_0(\mathfrak{s}(t)),y_0(\mathfrak{s}(t))]^\text T,\\
\boldsymbol{p}_{c0}
(t)&=\boldsymbol{p}_0(t)+[L_\text{rear}\cos\theta_0(t),L_\text{rear}\sin\theta_0(t)]^\text T,\\
\boldsymbol{p}_{ci}(t)&=\boldsymbol{p}_{i}(t)=\boldsymbol{p}_{i-1}(t)-\boldsymbol{R}_i(t)[L_i,0]^\text T,\\
\boldsymbol{R}_i(t)&=\left[\begin{matrix}
\cos\theta_i(t) & -\sin\theta_i(t)\\
\sin\theta_i(t) & \cos\theta_i(t)
\end{matrix}\right],\quad\quad\quad i=1,...,\text N.
\end{align}
$L_\text{rear}$ is the distance between $\boldsymbol{p}_0$ and the center of the tractor. The inequality in Conditions (\ref{con:safe}) is also taken element-wise. Conditions (\ref{con:safe}) are also include vehicle-to-vehicle collision avoidance. We impose the constraint that the centers $\boldsymbol{p}_{ci},i=0,1,...,\text N$ of each vehicle of the robot must be greater than a threshold distance from each other.

\subsection{Problem Simplification and Solving}
\label{subsec:Problem Simplification and Solving}
Observing Eq.(\ref{con:mincop})$\sim$Eq.(\ref{con:mincot}), we can see that the value of $\boldsymbol{C}$ is related to $\boldsymbol{P},\boldsymbol{S},\boldsymbol{\Theta},\boldsymbol{e},T_f$. Inspired by the work\cite{minco}, we add the constraint of the trajectories having zero second-order derivatives at the start and termination time to ensure that the matrices $\boldsymbol{M}_p,\boldsymbol{M}_\mathfrak{s},\boldsymbol{M}_\theta$ are invertible. Thus using the method proposed in the work\cite{minco} to convert the optimization variables $\{\boldsymbol{C},\boldsymbol{e},T_f\}$ to $\{\boldsymbol{P},\boldsymbol{S},\boldsymbol{\Theta},\boldsymbol{e},T_f\}$, we can eliminate the equation constraints (\ref{con:mincop})$\sim$(\ref{con:mincot}) while reducing the dimension of the optimization variables. 

To deal with the constraints $T_f>0,\mathfrak{s}(t)\ge0$, we refer to the differential homogeneous mapping\cite{diffmorphi}, using the following computationally convenient $C^2$ function to convert each element of $\boldsymbol{S}\in\mathbb{R}_{>0}^{\text{M}\times1}$ and  $T_f\in\mathbb{R}_{>0}$ to $\boldsymbol{\mathfrak{S}}\in\mathbb{R}^{\text{M}\times1}$ and $\tau_f\in\mathbb{R}$, respectively:
\begin{align}
\text L_{c2}(x)=\left\{\begin{matrix}
1-\sqrt{2x^{-1}-1} & 0< x\le1\\
\sqrt{2x-1}-1 & x>1
\end{matrix}\right.\quad.
\end{align}
With the constraint $\dot{\mathfrak{s}}(t)\ge0$, we can then ensure $\mathfrak{s}(t)\ge0$. Also, to guarantee that constraint (\ref{con:dyn_begin}) is satisfied when $t = T_f$, we use an inverse sigmoid-like function to convert $\boldsymbol{\theta}^\text{fina}$ to $\boldsymbol{\vartheta}_d^\text{fina}=[\vartheta_{d1}^\text{fina},\vartheta_{d2}^\text{fina},...,\vartheta_{d\text{N}}^\text{fina}]$,the corresponding equations are as follows: 
\begin{align}
\theta_{di}^\text{fina}&=\theta_{i-1}^\text{fina}-\theta_i^\text{fina}&i=1,2,...,\text N,\\
\vartheta_{di}^\text{fina}&=\text{L}_{c2}\left(\frac{\delta\theta_\text{max}+\theta_{di}^\text{fina}}{\delta\theta_\text{max}-\theta_{di}^\text{fina}}\right)&i=1,2,...,N.
\end{align}

For the remaining kinematic constraints (\ref{con:eqc}) and other inequality constraints, we discretize each piece of the duration $T_{p}$ as $\text K$ time stamps $\tilde{t}_{p}=(p/\text K)\cdot T_{p},p=0,1,...,K-1$, and impose the constraints on these time stamps.

Then, we use Powell-Hestenes-Rockafellar Augmented Lagrangian method\cite{alm} (PHR-ALM) to solve the simplified problem, which is an effective method for solving large-scale problems and those with complex constraints. To give readers a more accurate understanding of our work, we give a proof of convergence of this algorithm in the Appendix \ref{appendix} and provide a quantitative analysis to show the degree to which constraints are satisfied.

Inspired by other works on robot trajectory optimization problems that also need to deal with equation constraints\cite{uneven,payload}, we set the trajectory of trailers have more pieces
(i.e. $\Omega > \text{M}$) to make it easier for the optimizer to converge and better fit the kinematic constraints (\ref{con:eqc}). Based on engineering experience, we found that the solver can achieve good efficiency and success when $\Omega = \text{ceil}(2.0\times\text{M})$, where function $\text{ceil}(x)$ serves to take the smallest integer not less than $x$.

\begin{algorithm}
    \caption{SE(2)-MHA: Fast multi-terminal path searching
    based on Hybrid-A* for tractor-trailers}
    \label{alg:path searching}
    \KwIn{grid map $\mathcal{M}$, start state $\boldsymbol{s}_\text{start}\in\mathbb{R}\times SO(2)^\text{N+1}$, target region $\textbf{Info}_e$, $d_\text{shoot}>0$, weights $\boldsymbol{w}_g,w_l,w_e$.}
    \KwOut{the best path $\mathcal{P}$}
    \Begin
    {
        $\mathcal{O},\mathcal{C},\mathcal{E}\leftarrow\varnothing,\mathcal{T}\leftarrow \textnormal{GetEnds}(\textbf{Info}_e),l=+\infty$\;
        $n_s\leftarrow\mathcal{O}.\textnormal{AddStart}(\boldsymbol{s}_\text{start}),\mathcal{E}.\textnormal{Insert}(n_s)$\;
        \While{$\neg\ \mathcal{O}.\textnormal{Empty}()$}
        {
            $n_c\leftarrow\mathcal{O}.\textnormal{Pop}(),\mathcal{C}.\textnormal{Insert}(n_c),\mathcal{E}.\textnormal{Insert}(n_c)$\;
            \For{\textbf{each} $\tau\in\mathcal{T}$}
            {
                \If{\textnormal{HasPath}($\tau$)}
                {
                    \textbf{continue}\;
                }
                \If{$\textnormal{ReachEnd}(n_c, \tau)\vee(\textnormal{Dist}(n_c, \tau)<d_\text{shoot}\wedge\textnormal{ShootEnd}(n_c, \tau))$}
                {
                    $\mathcal{P}_a\leftarrow\textnormal{GetFullPath}(n_c)$\;
                    $l_a=\textnormal{CalcCost}(\mathcal{P}_a,\textbf{Info}_e,w_l,w_e)$\;
                    \If{$l_a<l$}
                    {
                        $l=l_a,\mathcal{P}=\mathcal{P}_a$\;
                    }
                }
            }
            \If{\textnormal{AllPathFound()}}
            {
                \textbf{return} $\mathcal{P}$.
            }
            \For{\textbf{each} $c \in \textnormal{GetInputs}()$}
            {
                $\boldsymbol{s}_c\leftarrow\textnormal{StateTrans}(n_c,c)$\;
                \If{$\neg\ \textnormal{InMap}(\boldsymbol{s}_c,\mathcal{M})$}
                {
                    \textbf{continue}\;
                }
                $n_t\leftarrow\mathcal{E}.\textnormal{Find}(\boldsymbol{s}_c)$\;
                \uIf{\textnormal{(}$n_t$ \textbf{is} $\textnormal{Null}\vee\neg\ \mathcal{C}.\textnormal{Find}(n_t))\wedge$\\$\textnormal{NoCollision}(n_c,c)$}
                {
                    $g_t\leftarrow n_c.g+\textnormal{GetCost}(n_c,c,\boldsymbol{w}_g)$\;
                    \uIf{$\neg\ \mathcal{O}.\textnormal{Find}(n_t)$}
                    {
                        $\mathcal{O}.\textnormal{SetAndInsert}(n_t),\mathcal{E}.\textnormal{Insert}(n_t)$\;
                    }
                    \textbf{else }\If{$g_t>n_t.g$}
                    {
                        $n_t.g\leftarrow g_t,n_t.parent\leftarrow n_c$\;
                        $n_t.f\leftarrow g_t+\textnormal{GetHeu}(n_t,\textbf{Info}_e)$\;
                    }
                }
            }
        }
        \textbf{return} $\varnothing$.
    }
\end{algorithm}

\begin{figure}[t]
    \centering
    \includegraphics[width=1.0\columnwidth]{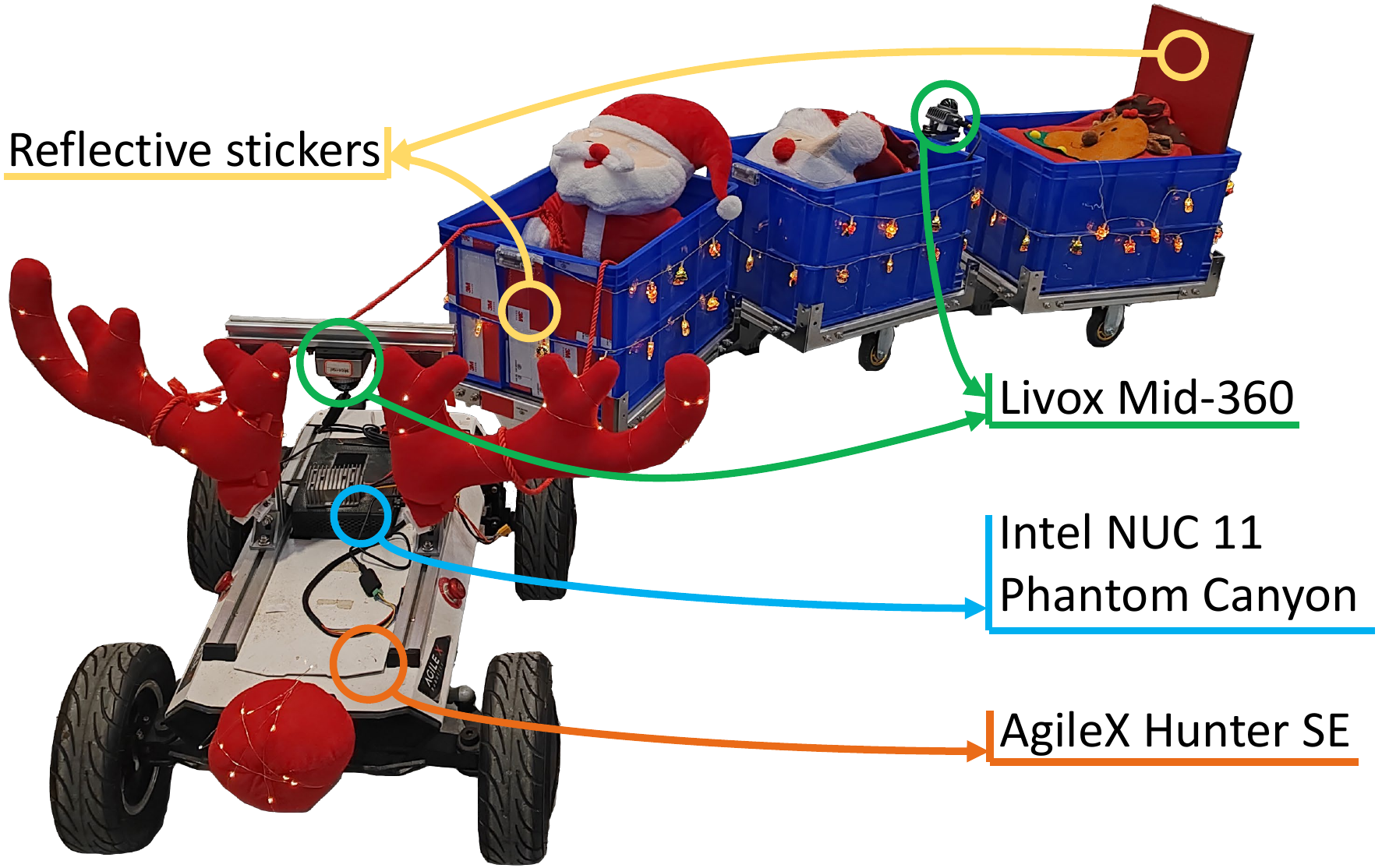}
	\caption{Tractor-trailer robot. It consists of three DIY trailers and an AgileX Hunter SE as the tractor.}\label{fig:hardware}
\end{figure}

\begin{figure}[t]
    \centering
    \includegraphics[width=1.0\columnwidth]{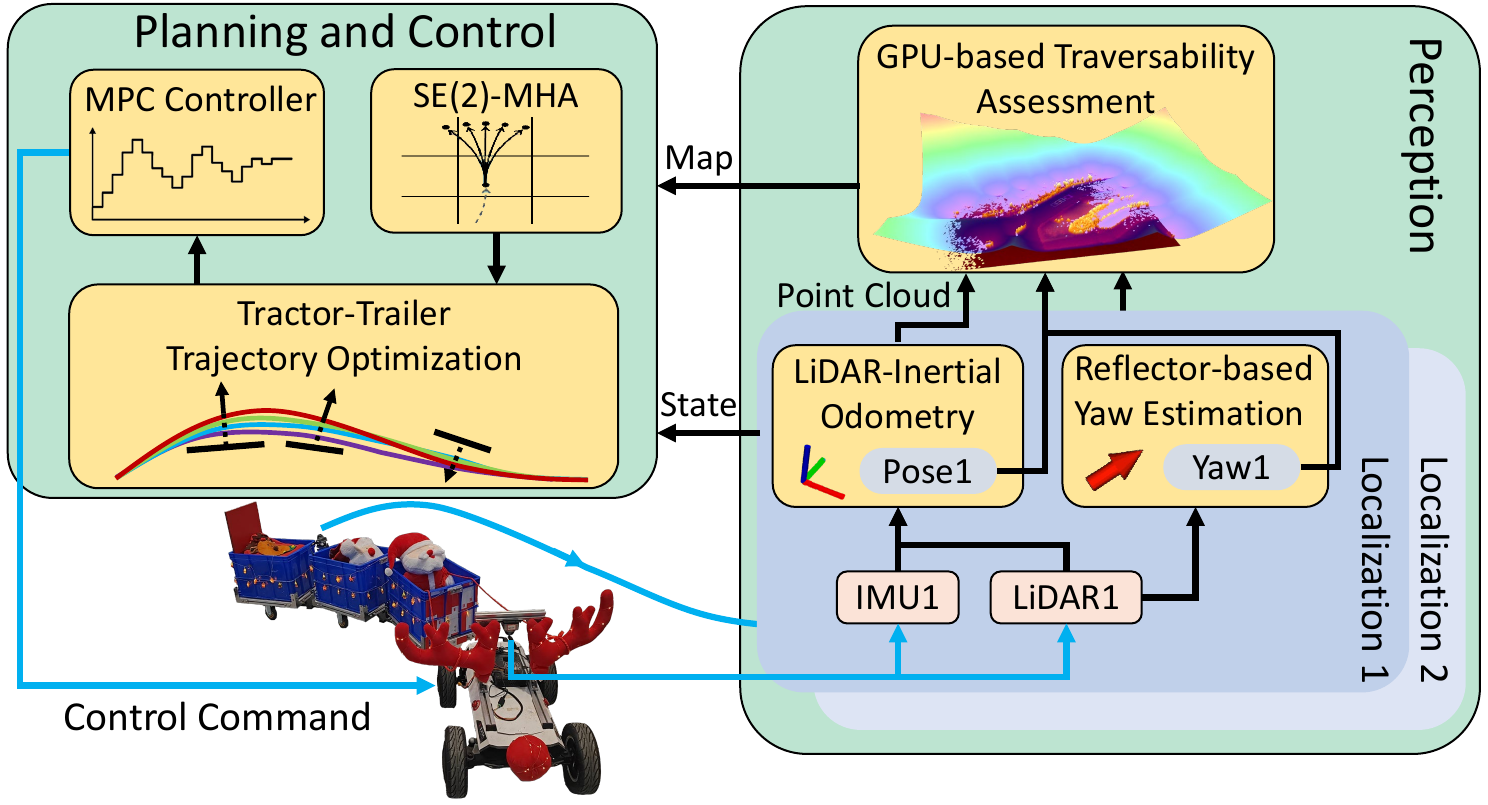}
	\caption{Software system of the robot. The perception module includes the global localization of the robot, its own state estimation, and traversability assessment. The planning and control module includes the proposed planning pipeline and trajectory tracking controller.}\label{fig:software}
\end{figure}

\subsection{Initial Value Acquisition}
\label{subsec:Initial Value Acquisition}
There are many methods that can be used to obtain initial values for trajectory optimization. Work\cite{libai_astar} uses the A* algorithm search directly for all vehicles of the robot to obtain the path quickly, but since this path is kinematically infeasible, the optimization will take longer to converge to a good solution. Work\cite{libai_pcoc} proposed an extended Hybrid-A* algorithm\cite{hybridAs} that searches the trajectory in the space $\mathbb{R}\times SO(2)^\text{N+1}$ including the position of the tractor and yaw angles of all vehicles, which can obtain a better initial value. However, due to the large search space, it usually takes much more time, which leads to the whole system not meeting the requirement of real-time replanning in unknown environments for avoiding just-appearing obstacles.

To balance the quality and efficiency, we propose a \textbf{m}ulti-terminal path search method based on \textbf{H}ybrid-\textbf{A}*\cite{hybridAs} SE(2)-MHA, which searches only in $SE(2)$ space for the tractor and uses an approximate integration method similar to that in work~\cite{libai_pcoc} to obtain the path and terminal position of the trailers in searched paths corresponding to each terminal. Finally, we evaluate the quality of the paths using the weighted cost of the path length and the distance of each vehicle from the target region, choosing the one with the smallest value as the initial value for trajectory optimization. Specifically, assuming the length of the tractor path is $l_\text{tra}$ and the distance between the terminal position of $i$-th trailer and the target convex area is $d_{it}, i=1, 2, \text N$, the cost of the path is $\text{Cost}_{traj}=w_\text{r2}l_\text{tra}+w_\text{trailer}\sum_{i=1}^\text{N}d_{it}$, where $w_\text{r2}$ and $w_\text{trailer}$ are the weights. In our algorithm, $d_{it}$ will be set to zero if the terminal position of the $i$-th trailer is within the target convex area. The pseudo code of the proposed method is given in Algorithm \ref{alg:path searching}.

In Algorithm \ref{alg:path searching}, $\mathcal{O}$ and $\mathcal{C}$ refer to the open and closed set respectively. While $\mathcal{E}$ denotes is used to store the expanded nodes for pruning to speed up the process of searching. $\mathcal{T}$ includes different end states, each state $\{\boldsymbol{p}_\lambda^e,\theta_\lambda^e\}$ is determined by the midpoint $\boldsymbol{p}_{\lambda c}$ of each edge of the convex polygon and the length $L_\text{f}$ of the tractor, $\boldsymbol{p}_\lambda^e=\boldsymbol{p}_{\lambda c}-(L_\text{rear}+0.5L_\text{f})\textbf{n}_\lambda,\theta_\lambda^e=\text{atan2}(\textbf{n}_\lambda(1),\textbf{n}_\lambda(0))$.To further accelerate the process, we also adopt Dubins Curve\cite{dubins} to shoot the end states for earlier termination when the distance between the current node and the end state is less than a threshold $d_\text{shoot}$. Once we get a feasible path for the tractor, $v_0$ and the change in $\theta_0$ between two neighboring states in the path will be estimated. We then approximate the states of the trailers by discretizing Eq.(\ref{eq:kin_cstr}). The state transfer equations Eq.(\ref{eq:kin_begin}),(\ref{eq:kin_y}) and (\ref{eq:kin_th}) are also be discretized for expanding the state with the discretized $v_0$ and $\delta$ as control inputs. In this work, we use a weighted summation of different values as the cost function $g$, including the distance of expansion, the magnitude of the change in yaw angle $\theta_0$, and the accumulation of control efforts with respect to time. For the heuristic function $h$, we chose the shortest Euclidean distance to the center of the target region.


\begin{figure}[t]
    \centering
    \includegraphics[width=1.0\columnwidth]{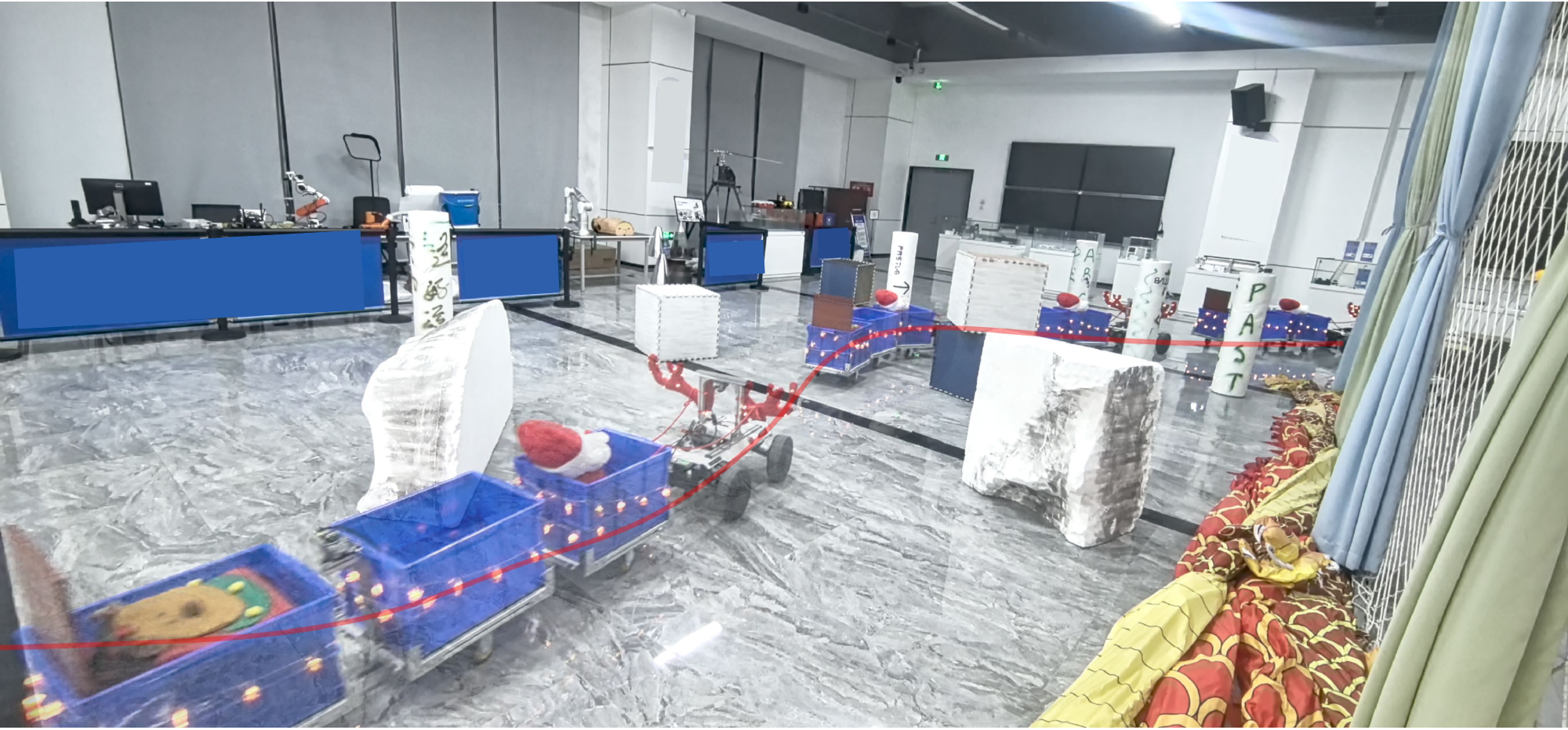}
	\caption{Part of the indoor experimental area. Tractor-trailer robot is traversing complex indoor environment.}\label{fig:indoor}
\end{figure}

\begin{table}[t]
	\small
	\centering
	\renewcommand\arraystretch{1.5}
	\caption{\label{tab:real_dyn} Statistics in Real-world Experiments}
	\begin{tabular}{llccc}
\toprule
Statistics                                           & Exp.                                             & Mean   & Max   & STD.   \\ \toprule
\multirow{3}{*}{$\hat{v}_0$ ($m/s$)}                 & Indoor                                           & 1.03   & 1.20  & 0.215  \\
                                                     & Outdoor (1\textsuperscript{st}) & 1.05   & 1.20  & 0.188  \\
                                                     & Outdoor (2\textsuperscript{nd}) & 1.07   & 1.20  & 0.184  \\ \hline
\multirow{3}{*}{$\hat{a}$ ($m/s^2$)}                 & Indoor                                           & 0.0651 & 0.649 & 0.0685 \\
                                                     & Outdoor (1\textsuperscript{st}) & 0.0758 & 0.625 & 0.0758 \\
                                                     & Outdoor (2\textsuperscript{nd}) & 0.0878 & 0.739 & 0.0883 \\ \hline
\multirow{3}{*}{$\hat{a}_\text{lat}$ ($m/s^2$)}      & Indoor                                           & 0.117  & 0.754 & 0.102  \\
                                                     & Outdoor (1\textsuperscript{st}) & 0.139  & 0.641 & 0.111  \\
                                                     & Outdoor (2\textsuperscript{nd}) & 0.157  & 1.16  & 0.125  \\ \hline
\multirow{3}{*}{$\hat{\delta\theta}$ ($\text{rad}$)} & Indoor                                           & 0.137  & 0.427 & 0.0930 \\
                                                     & Outdoor (1\textsuperscript{st}) & 0.110  & 0.442 & 0.0887 \\
                                                     & Outdoor (2\textsuperscript{nd}) & 0.0899 & 0.452 & 0.0828 \\ \hline
\multirow{3}{*}{$\text{Tracking Error}$ ($m$)}       & Indoor                                           & 0.115  & 0.172 & 0.0226 \\
                                                     & Outdoor (1\textsuperscript{st}) & 0.115  & 0.248 & 0.0270 \\
                                                     & Outdoor (2\textsuperscript{nd}) & 0.114  & 0.242 & 0.0211 \\ \bottomrule
\end{tabular}
\end{table}

\begin{figure*}[t]
		\centering
		{\includegraphics[width=1.0\linewidth]{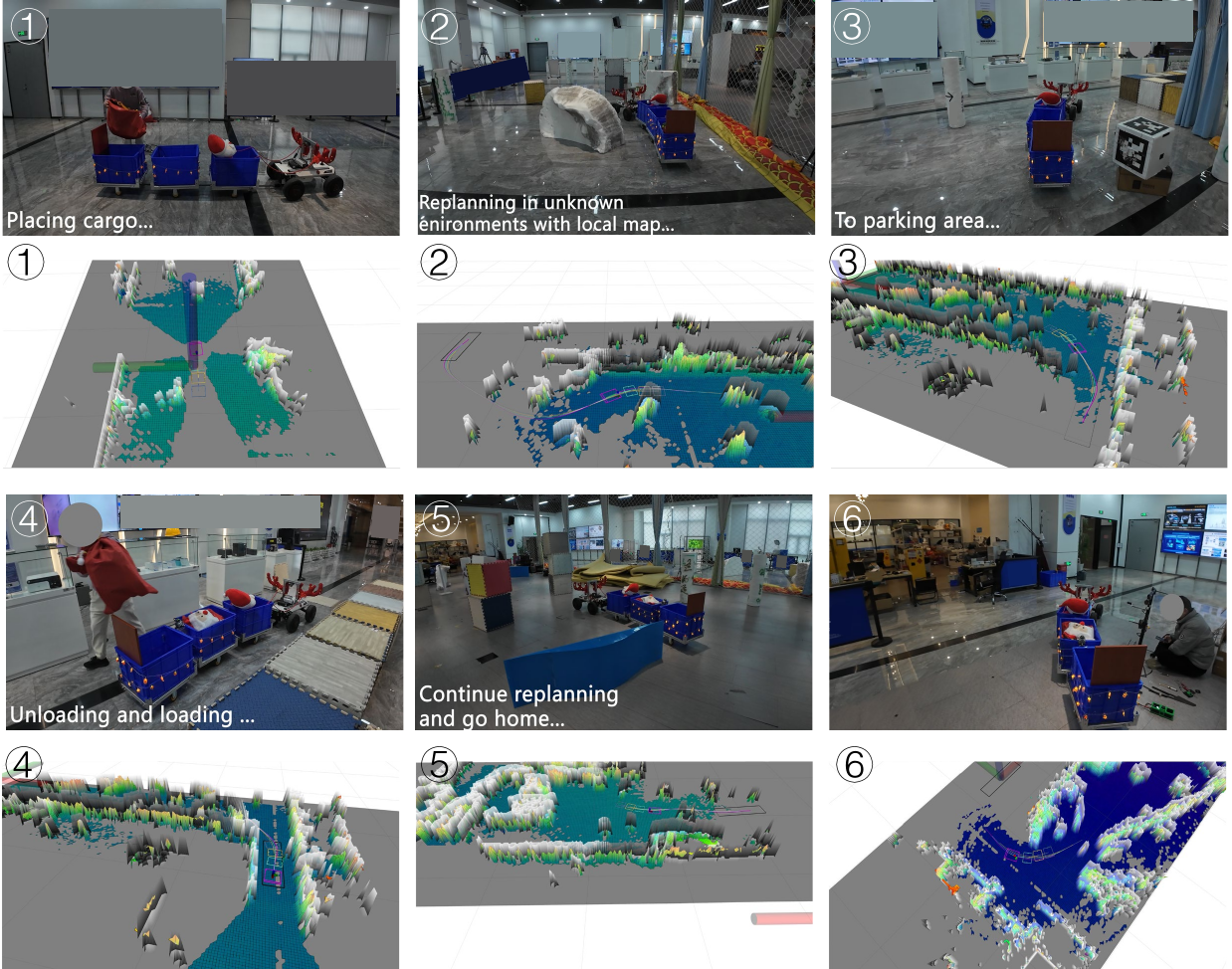}}
		\caption{The whole process of the indoor experiment. Different numerical markers indicate third-person viewpoints or corresponding views from Rviz tacken at different moments. In Rviz, the coloured part indicates the robocentric elevation map, and the semi-transparent black and white part indicates the traversability map, with higher indicating less traversable.}
		\label{fig:indoor_all}
\end{figure*}

\section{Results}
\label{sec:Results}
\subsection{Implementation details}
\label{subsec:Implementation details}
To validate the performance of our method in real-world applications, we deploy it on a tractor-trailer robot, as shown in Fig. \ref{fig:hardware}. The robot is driven by an Ackermann chassis and carries three easily removable passive trailers. We use hinges that connect the trailer to the tractor and the trailers to the trailers.

Fig.~\ref{fig:software} illustrates the software system of the robot. In the perception module, we use the combination of LiDARs and reflective stickers\cite{swarm-lio,swarm-track} to acquire the state of the robot, which includes the pose of the tractor and the yaw angle of each trailer. Specifically, the reflective stickers of the latter vehicle provides a high-intensity point cloud for the former LiDAR-installed vehicle. Then the relative pose of the two vehicles can be obtained by filtering this part of the point cloud and computing the normal vector of the fitted plane. We run LiDAR-inertia Odometry (LIO) on two LiDARs separately. Using the initial point cloud map constructed by the LiDAR on the tractor as initial map of the LIO on the trailer, we can align the coordinate system of these two LIOs. In this work, we adopt FAST-LIO2\cite{fast-lio2} as the LIO algorithm.

The registered point clouds from the tractor and states of the robot are further used to construct elevation map\cite{ele_map} and perform traversability assessment. For real-time requirements, we remove dynamic obstacles using ray casting and compute the covariance matrix of the point cloud in parallel with the help of GPU\cite{ele_gpu}. Further, still in parallel, we patch unseen grids using the nearest-neighbour method, computing the normal direction of the fitted plane and the approximate curvature\cite{terrain_curvature} for each grid. We obtain the final robocentric occupancy grid map by setting thresholds for traversability. Subsequently, the resulting map is delivered to the planning and control module, where the SDF will be computed by an efficient $O(n)$ algorithm\cite{esdf}.

\begin{figure*}[t]
		\centering
		{\includegraphics[width=1.0\linewidth]{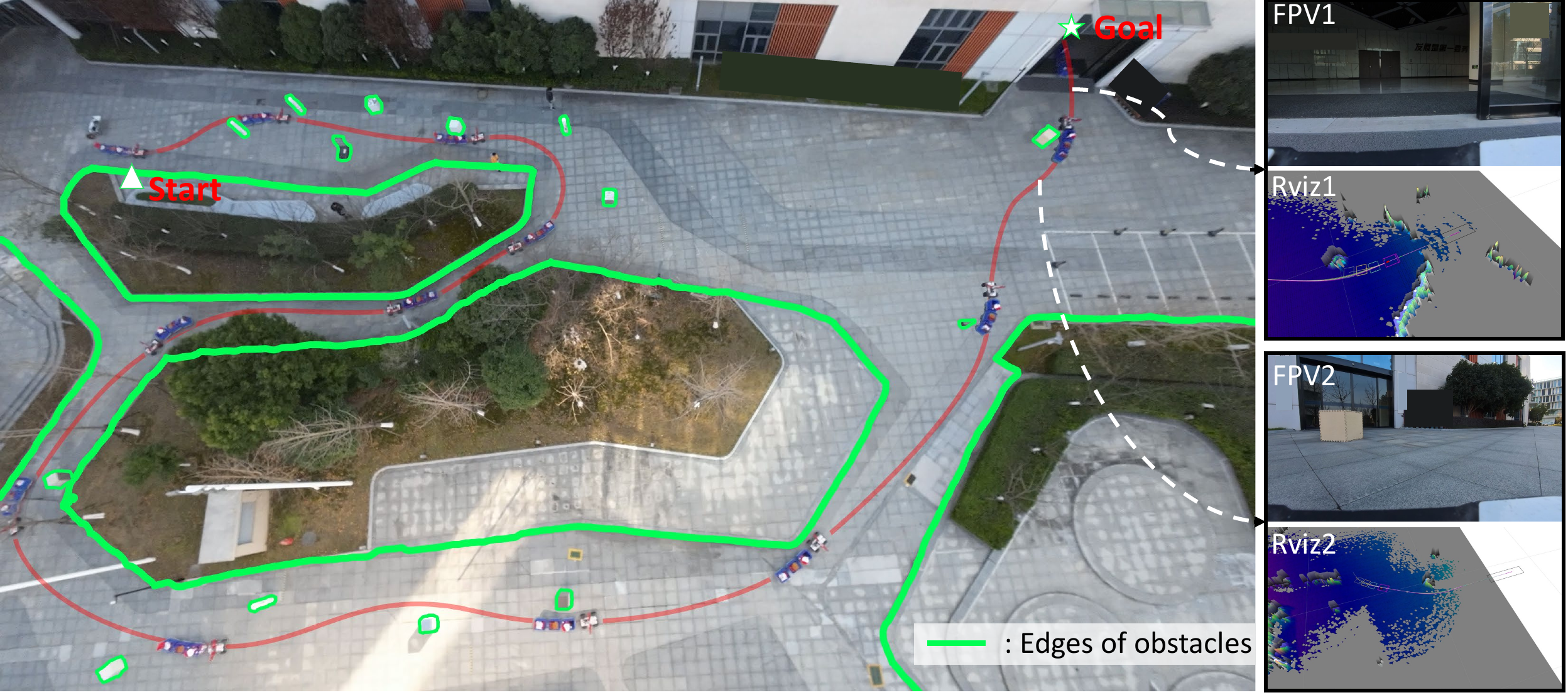}}
		\caption{The second scene of outdoor experiments. The tractor-trailer robot autonomously traverses the park using the proposed planning algorithm to enter the interior of the building through an automatic door.}
		\label{fig:outdoor2}
\end{figure*}

A replanning process is activated when a set timer is triggered or the target region is updated. It will call the SE(2)-MHA algorithm with the current or predicted state of the robot as the starting state and perform trajectory optimization. A MPC controller will receive the optimized trajectory to compute control commands consisting of speed and steering angle, which utilizes the position of the tractor and the yaw angles of all vehicles of the robot as state. The objective function includes the norm of the error between the states and the reference states, as well as the norm of the control inputs. We use Casadi\cite{casadi} to construct the MPC problem, solving it by Sequential Quadratic Programming.

We also build some simulation environments based on ROS\footnote{\url{https://www.ros.org/}} to testify the effectiveness of our method in various scenarios and conduct comparative experiments with other algorithms, as detailed in Sec.\ref{subsec:Cases Study} and Sec.\ref{subsec:Benchmark Comparisons}. All simulations are run on Ubuntu 20.04 with an Intel i9-14900 CPU and a GeForce RTX 4090D GPU.

\subsection{Real-World Experiments}
\label{subsec:Real-World Experiments}
We present several experiments in both indoor and outdoor cluttered environments. In the indoor experiment, we set up the tractor-trailer robot to transport, load and unload cargo to demonstrate the possible real-life applications of the proposed navigation system, as illustrated in the screenshots of Fig.~\ref{fig:indoor_all}. At the beginning, the cargo is placed into one trailer. Receiving the target loading and unloading area, the robot starts to continuously replan the trajectory using the real-time updated map to avoid unknown obstacles. After traversing the complex environment, as shown in Fig.~\ref{fig:indoor}, and arriving at the parking area where its cargo is loaded and unloaded, the robot continues to replan and returns to the starting area.

To validate the effectiveness of proposed framework in more scenarios, we conduct two outdoor experiments. The first outdoor scene is the aisle by the non-motorized parking area, as shown in Fig.~\ref{fig:head_image}. This environment is filled with different kinds of obstacles including cube-shaped, cylindrical, carts, cars, bicycles, e-bikes etc. The robot needs to constantly replan to avoid obstacles and traverse this narrow passages. The second outdoor experiment is illustrated in Fig.~\ref{fig:outdoor2}. In this scenario, the robot is required to load some cargo and drive into a building by approaching predetermined waypoints. Similar to indoor experiments, the robot perceives the world through the local traversability map updated in real time, and continuously replans to avoid unknown obstacles along the way. Eventually, it enters the building by triggering an automatic door. 

In the real-world experiments, limitation of longitudinal velocity, longitude acceleration, latitude acceleration, curvature and difference in yaw angle between adjacent vehicles are set to $v_\text{mlon}=1.2m/s$, $a_\text{mlon}=0.8m/s^2$, $a_\text{mlat}=1.2m/s^2$, $\kappa_\text{max}=0.6m^{-1}$ and $\delta\theta_\text{max}=0.9\text{ rad}$, respectively. Meanwhile, we set the time weight $\rho_t=10.0$ to ensure the aggressiveness of the trajectory. 

Tab.~\ref{tab:real_dyn} shows the dynamic evaluation metrics during the real-world experiments for the robot, where $\hat{v}_0$ is the estimated longitudinal velocity of the tractor, given by the wheel tachometer. $\hat{a}$ and $\hat{a}_\text{lat}$ are measured longitude acceleration and latitude acceleration, respectively, obtained from the IMU accompanying the LiDAR. $\hat{\delta\theta}$ is the largest estimated difference between the yaw angles of two neighboring vehicles, read from the estimated robot state. Tracking Error is the distance between the current position and the reference position of the tractor. By integrating $\hat{v}_0$ with respect to time, we calculate the total distance traveled by the tractor to be $36.178m$, $26.774m$ and $81.987m$ in the indoor, first and second outdoor experiments, respectively. 

\begin{table}[t]
        \fontsize{8pt}{10pt}\selectfont
	\centering
	\renewcommand\arraystretch{1.2}
	\caption{\label{tab:param_cs} Key Parameter Settings for Simulation in Parking Lot. M.P. represents the proposed method. M.O. represents OCP-STC\cite{libai_corridor}. The specific names of abbreviations $\text{m.s., mu.s., b.p., l.s., m.c.t}$ are ``mem size", ``mu strategy", ``bound push", ``linear solver", ``max cpu time", respectively.}
        \begin{tabular}{c|l|l|c}
\toprule
Type                                                                           & \multicolumn{1}{c|}{Para.} & \multicolumn{1}{c|}{Description}                                                                                              & Setting     \\ \midrule
\multirow{8}{*}{\begin{tabular}[c]{@{}c@{}}M.P.\end{tabular}} & $L_q$                       & \begin{tabular}[c]{@{}l@{}}Intervals for taking state points from the\\ front-end path as optimization variables\end{tabular} & 5 m         \\ \cline{2-4} 
                                                                               & $K$                         & \begin{tabular}[c]{@{}l@{}}The number of points sampled to impose\\ constraints in each polynomial trajectory\end{tabular}    & 16          \\ \cline{2-4} 
                                                                               & m.s.                    & \begin{tabular}[c]{@{}l@{}}The number of corrections to approximate\\ the inverse hessian matrix in L-BFGS\end{tabular}       & 64          \\ \cline{2-4} 
                                                                               & $\delta_\text{inner}$       & Desired convergence tolerance for L-BFGS                                                                                      & 1.0$^{-4}$  \\ \cline{2-4} 
                                                                               & past                        & Distance for delta-based convergence test                                                                                     & 3           \\ \cline{2-4} 
                                                                               & $N_\text{iter}$            & Maximum iteration number of L-BGFS                                                                                            & 10000       \\ \cline{2-4} 
                                                                               & $\epsilon_\text{cond}$      & Desired convergence tolerance for ALM                                                                                         & 1.0$^{-1}$  \\ \cline{2-4} 
                                                                               & $N_\text{out}$             & Maximum iteration number of ALM                                                                                               & 100         \\ \hline
\multirow{5}{*}{M.O.}                                                   & $L_{fe}$                    & \begin{tabular}[c]{@{}l@{}}Intervals for initializing finite\\elements from the front-end path\end{tabular}                    & 0.3 m         \\ \cline{2-4} 
                                                                               & mu.s.                 & Update strategy for barrier parameter                                                                                         & adaptive    \\ \cline{2-4} 
                                                                               & b.p.                & \begin{tabular}[c]{@{}l@{}}Desired minimum absolute distance from \\ the initial point to bound\end{tabular}                  & 1.0e$^{-8}$ \\ \cline{2-4} 
                                                                               & l.s.               & The core linear solver for Ipopt                                                                                              & MA97        \\ \cline{2-4} 
                                                                               & m.c.t                & \begin{tabular}[c]{@{}l@{}}Limit on CPU seconds that Ipopt can \\ use to solve one problem\end{tabular}                       & 200.0 s     \\ \bottomrule
\end{tabular}
\end{table}

\begin{figure}[t]
    \centering
    \includegraphics[width=1.0\columnwidth]{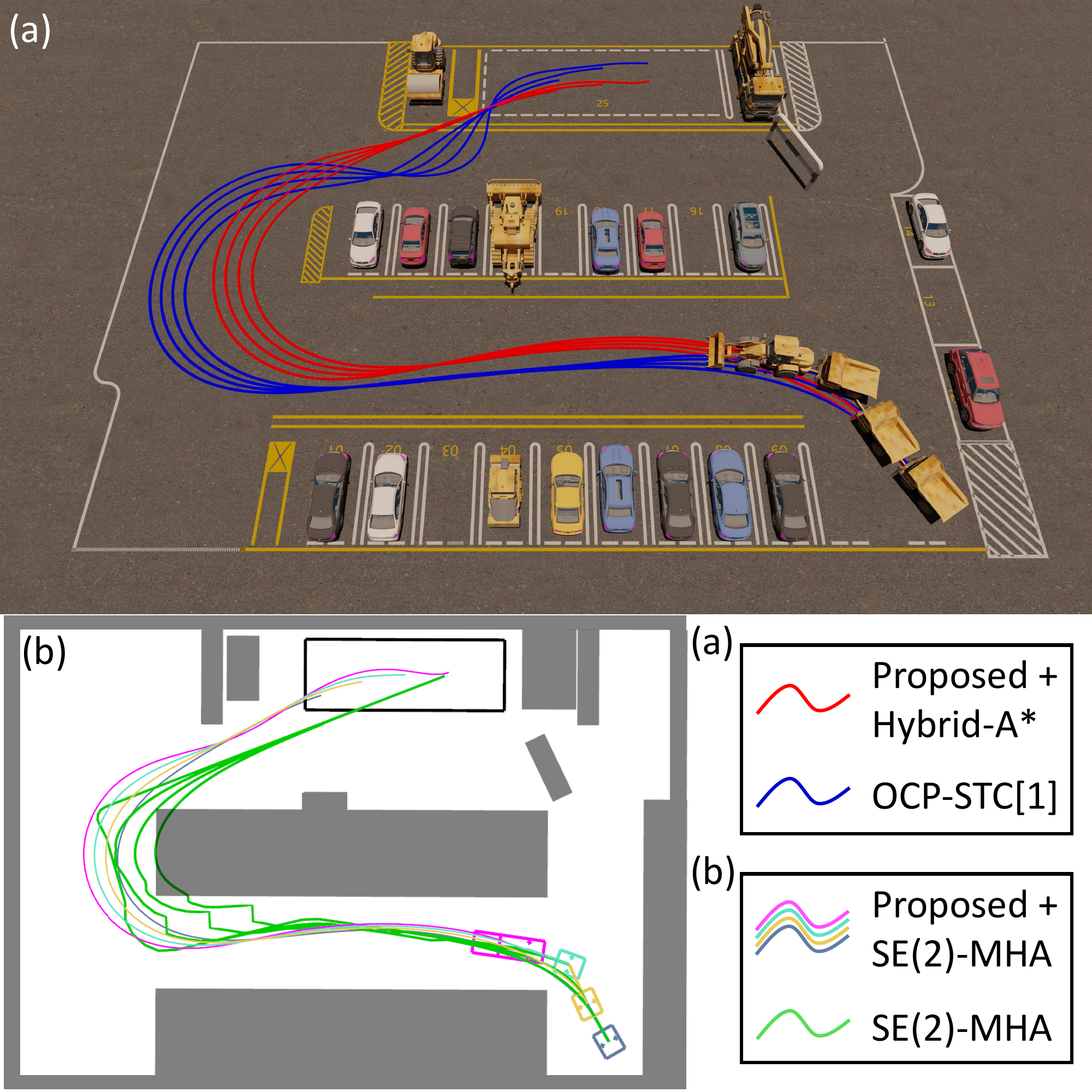}
	\caption{A bulldozer pulls three transport trucks into a parking space in a parking lot. Figure (a) visualizes this task and the trajectories generated using Hybrid-A* as front end. In this case, the distribution of the obstacles is shown by the gray area in Figure (b).}\label{fig:park}
\end{figure}

As we can see from Tab.~\ref{tab:real_dyn}, the robot maintains a high speed and small tracking error when traversing the environment. The small mean and variance of the latitude and longitude accelerations indicate that the robot can keep a high stability during driving despite the need for constantly replanning when facing unknown environments. In addition, the maximum value of each metric reveals that there is no violation of the constraints we set for the robot. More demonstrations can be found in the attached multimedia.

\begin{table*}[t]

\small
\centering
\renewcommand\arraystretch{1.5}
\caption{\label{tab:park} Statistics in a parking lot}
\begin{tabular}{cc|ccccc}
\hline
\begin{tabular}[c]{@{}c@{}}Front End\\Method\end{tabular} & \begin{tabular}[c]{@{}c@{}}Search\\Time (ms)\end{tabular} & \begin{tabular}[c]{@{}c@{}}Optimization\\Method\end{tabular} & \begin{tabular}[c]{@{}c@{}}Corridor Generation\\Time (ms)\end{tabular} & \begin{tabular}[c]{@{}c@{}}Optimization\\Time (ms)\end{tabular} & \begin{tabular}[c]{@{}c@{}}Planning Total\\Time (s)\end{tabular} & \begin{tabular}[c]{@{}c@{}}Trajectory\\Length (m)\end{tabular} \\ \hline
\multirow{2}{*}{Hybrid-A*}                                 & \multirow{2}{*}{1224}                                      & OCP-STC                                                       & 27.7                                                                     & 2892.5                                                           & 4.14                                                              & 100.7                                                           \\
                                                           &                                                            & Proposed                                                      & 0.0                                                                      & 156.36                                                           & 1.38                                                              & 86.75                                                           \\ \hline
\multirow{2}{*}{SE(2)-MHA}                                 & \multirow{2}{*}{178.7}                                     & OCP-STC                                                       & /                                                                        & /                                                                & /                                                                 & /                                                               \\
                                                           &                                                            & Proposed                                                      & 0.0                                                                      & 185.53                                                           & \textbf{0.364}                                                    & \textbf{86.66}                                                  \\ \hline
\end{tabular}
\end{table*}

\subsection{Cases Study}
\label{subsec:Cases Study}
In simulations, we first perform a qualitative and simple quantitative comparison for proposed method and OCP-STC\cite{libai_corridor} in a parking lot scenario with two different initial value acquisition methods, as shown in Fig.~\ref{fig:park}, where OCP-STC\cite{libai_corridor} is an efficient trajectory optimization method for tractor-trailer robots in unstructured environments, which uses Ipopt\cite{ipopt} as the solver and imposes obstacle avoidance constraints using safe travel corridors. In this case, we set $L_0=2.7m,\text{N}=3,L_1=L_2=L_3=4.0m.$ The width of the tractor and trailers $L_w=2.0m$. The constants related to kinematic constraints are set to $v_\text{mlon}=10.0m/s,a_\text{mlon}=5.0m/s^2,a_\text{mlat}=10m/s^2,\kappa_{\text{max}}=0.95m^{-1},\delta\theta_\text{max}=1.47\text{rad}$, respectively. As in the original paper\cite{libai_corridor}, we use AMPL\cite{ampl} to construct the optimization problem for OCP-STC\cite{libai_corridor}. Moreover, the list of key parameter settings for optimization algorithms is given in Table~\ref{tab:param_cs}. The results using modified Hybrid-A* proposed in the work\cite{libai_pcoc} as the initial value are shown in Fig.~\ref{fig:park}(a). Since the presence of narrow spaces and large turns in this scenario, it is difficult to provide good seeds from the front-end path for corridor generation for every vehicle in the robot, the OCP-STC\cite{libai_corridor} causes the robot to take bigger turns and does not make good use of the collision-free area at the edge of the parking space.

The results of the quantitative comparison are demonstrated in Table \ref{tab:park}. Since the SE(2)-MHA expands the states only in $SE(2)$ space for the tractor and does not perform collision check on the trailers, the time consumed by path search is reduced by almost an order of magnitude compared to Hybrid-A*. Although such an initial value leads to a longer optimization time, it reduces the total planning time. The corridor-based approach OCP-STC\cite{libai_corridor} requires completely collision-free initial values for corridor generation. Thus SE(2)-MHA cannot be used as an initial value. In addition, the corridors narrow the solution space of the problem, resulting in it not containing a better solution in this case, giving a longer trajectory length.

\begin{figure}[t]
    \centering
    \subcaptionbox{Trajectories in the simulation.}{\includegraphics[width=1.0\columnwidth]{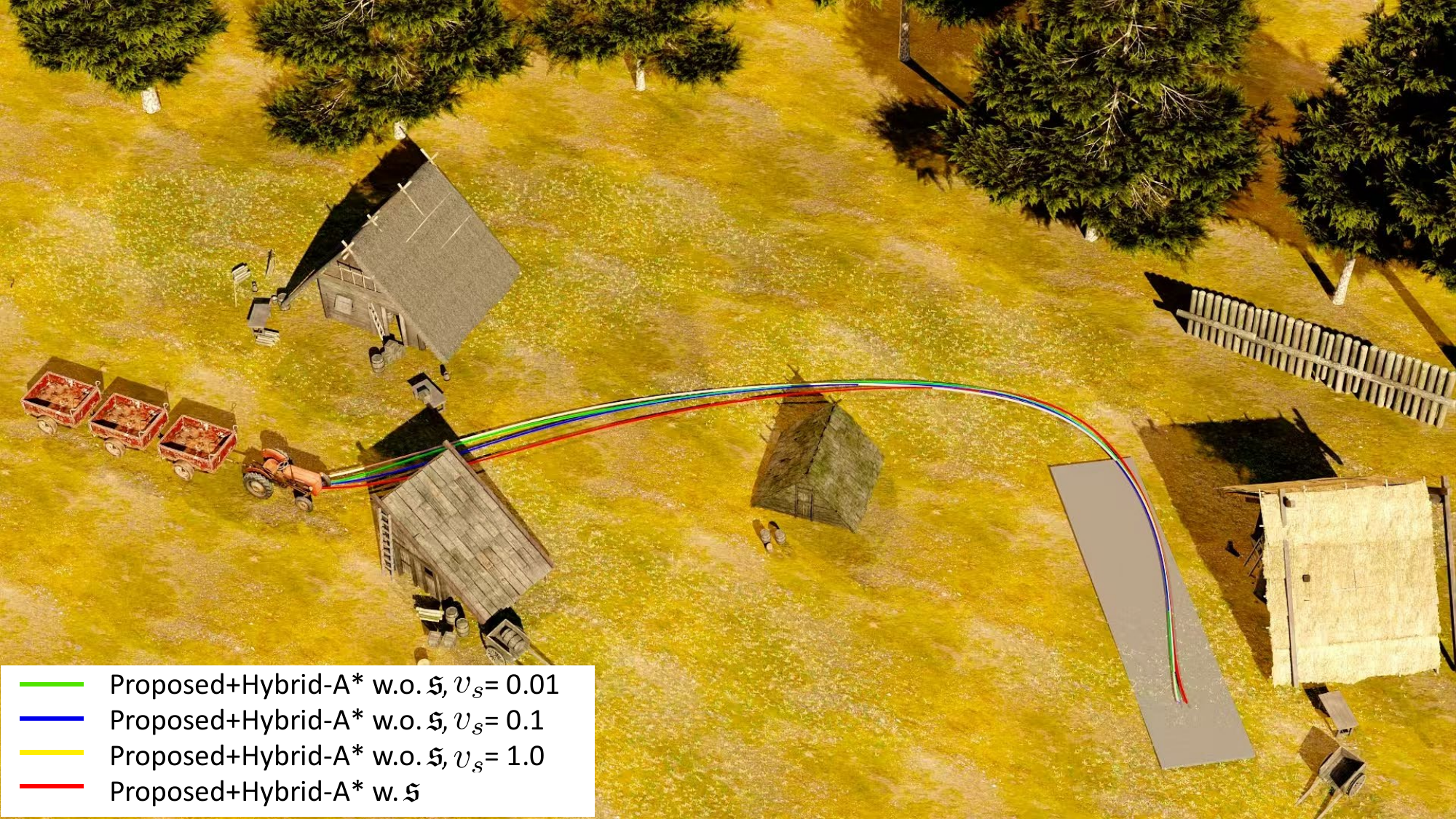}}
    \subcaptionbox{Speed, curvature, tracking error curves of optimized trajectories.}{\includegraphics[width=1.0\columnwidth]{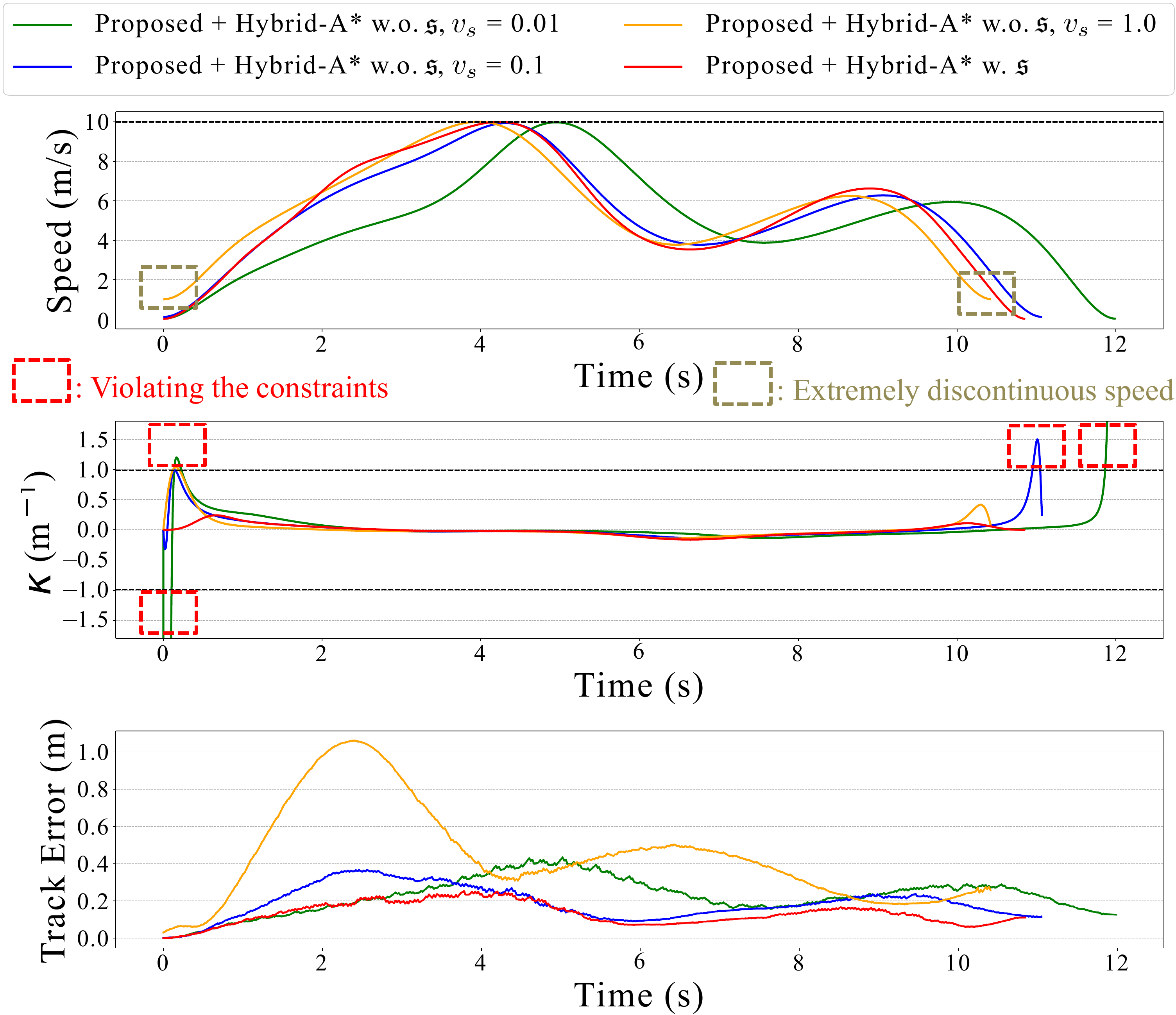}}
    \caption{The tractor pulls three transport trailers through the simulated village environment. The curves in Figure (a) represent the trajectories of the tractor.}
    \label{fig:village}
\end{figure}

The second case is an ablation experiment to demonstrate the importance of \textit{slackened arc length} in a simulated village, as shown in Fig.~\ref{fig:village}, where $v_s$ is the minimum speed constrained to avoid singularities. In this case, we set $L_0=2.6m,\text{N}=3,L_1=L_2=L_3=4.0m.$ The width of the tractor and trailers $L_w=2.8m$. The constants related to kinematic constraints are set to $v_\text{mlon}=10.0m/s,a_\text{mlon}=5.0m/s^2,a_\text{mlat}=2m/s^2,\kappa_{\text{max}}=0.99m^{-1},\delta\theta_\text{max}=1.47\text{rad}$, respectively. Moreover, the key parameters of the optimization algorithm are set the same as those given in Table~\ref{tab:param_cs} for methods with and without the introduction of slackened arc length $\mathfrak{s}$. The black dashed lines in Fig.~\ref{fig:village}(b) indicate the boundaries of the constraints. As can be seen in Fig.~\ref{fig:village}(b), the smaller the $v_s$, the more significant the effect of singularity on trajectory optimization. Although the smaller the $v_s$, the more the optimizer is able to constrain the maximum curvature of the robot, it will make the speed of the tractor mutation at the initial and end states more severe, which is not in accordance with the real-world situation, and may cause difficulties for accurate trajectory tracking. We introduce a model predictive control (MPC)-based controller\cite{mpc} to measure the tracking error in the simulation. In our paper, the tracking error of a tractor-trailer robot is defined as the average tracking error of all vehicles in the robot. The statistical results are shown in Table~\ref{tab:track_err_sim} and the curves in Fig.~\ref{fig:village}(b). Optimized trajectories without the use of $\mathfrak{s}$ lead to larger tracking errors due to abrupt changes in velocity or constraints violations during the initial and end phases of the trajectories. Especially when $v_s = 1.0$, extremely discontinuous velocities cause the controller to almost diverge at the beginning.

\begin{table}[t]
	\small
	\centering
	\renewcommand\arraystretch{1.2}
	\caption{\label{tab:track_err_sim} Tracking error statistics for ablation experiment.}
        \begin{tabular}{cccc}
        \toprule
        \multirow{2}{*}{Method}                                                           & Mean T.E.      & Max T.E.       & STD.            \\
                                                                                  & (m)            & (m)            & (m)             \\  \midrule
        \begin{tabular}[c]{@{}c@{}}Proposed w.o. $\mathfrak{s}$\\ $v_s=0.01$\end{tabular} & 0.227          & 0.434           & 0.0959          \\\hline
        \begin{tabular}[c]{@{}c@{}}Proposed w.o. $\mathfrak{s}$\\ $v_s=0.1$\end{tabular}  & 0.191          & 0.366          & 0.0868          \\\hline
        \begin{tabular}[c]{@{}c@{}}Proposed w.o. $\mathfrak{s}$\\ $v_s=1.0$\end{tabular}  & 0.443          & 1.06          & 0.259           \\\hline
        Proposed w. $\mathfrak{s}$                                                        & \textbf{0.133} & \textbf{0.254} & \textbf{0.0595} \\ \bottomrule
        \end{tabular}
\end{table}

\begin{figure}[t]
    \centering
    \includegraphics[width=1.0\columnwidth]{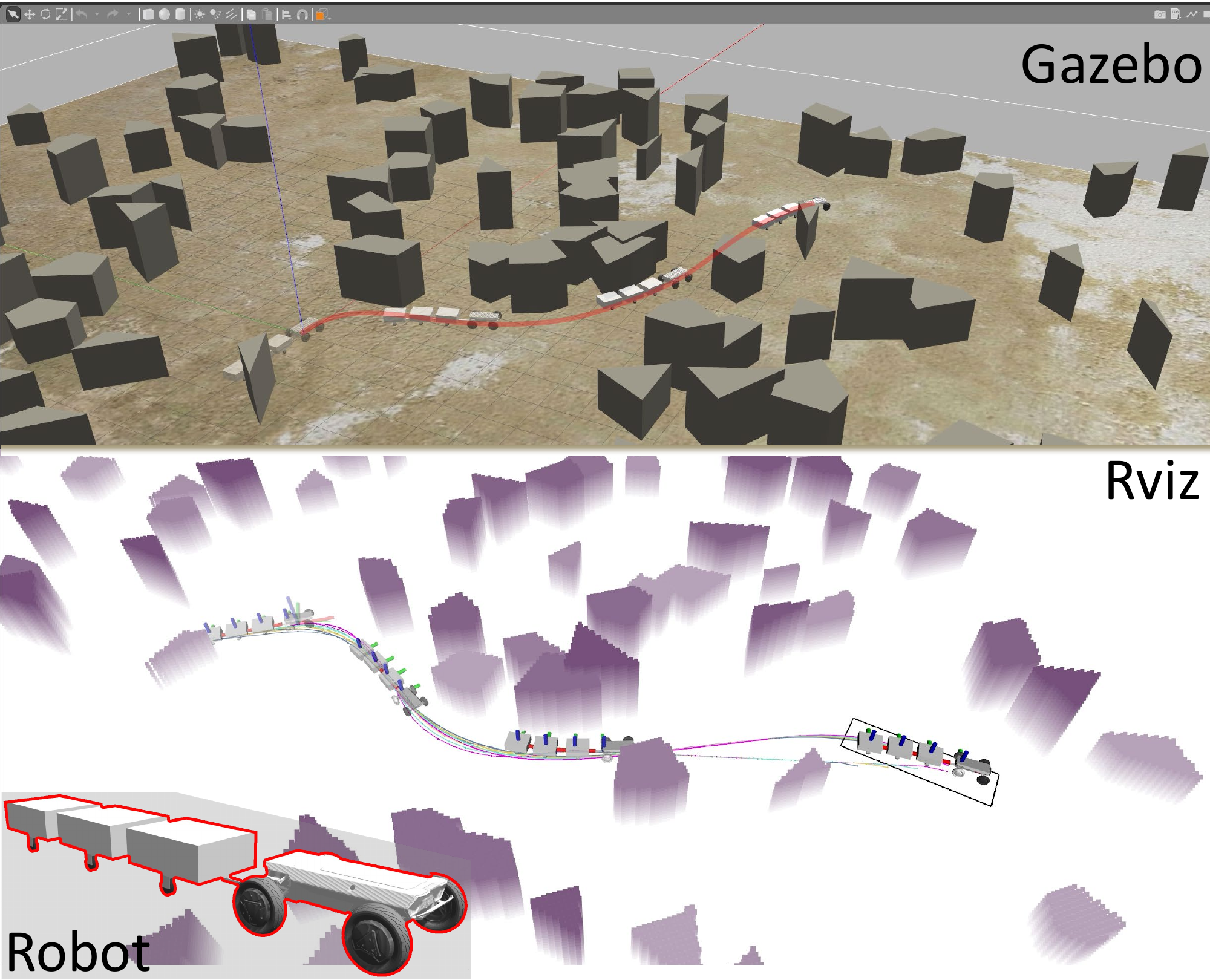}
    \caption{Physical simulation based on Gazebo. In Rviz, the purple parts indicate obstacles. The height of the point cloud from low to high is indicated by the color from white to purple.}\label{fig:gazebo}
\end{figure}

\subsection{Benchmark Comparisons}
\label{subsec:Benchmark Comparisons}
We carry out benchmark comparisons in a simulation environment of size $40m\times40m$ with random triangular, quadrilateral, and pentagonal obstacles as shown in Fig. \ref{fig:gazebo}. We randomly initialize the initial state and target region of the robot in this environment. Among them, the initial state of the robot includes the $SE(2)$ state of the tractor and the yaw angle of each trailer. We use a rectangle as the target region, given that parking areas in the real world are usually of this shape. We set the length and width of the tractor trailer be $0.6m$ and $0.4m$, respectively. The length and width of the trailer are both set to be $0.4m$, the maximum speed is $2.0m/s$, the maximum longitudinal acceleration is $2.0m/s^2$, the minimum angle at which jacknife occurs is $1.47$rad. The wheelbase length of the tractor is set to $0.5m$, and the maximum steering angle of it is $0.7$ rad. The parameters of all algorithms are finely tuned for the best performance. Moreover, the key parameters for optimization algorithms are consistent with those in Table~\ref{tab:param_cs}, except that considering the scene and robot size, we set $L_q=1.0$ m, $K=8$, and $L_{fe}=0.1m$. As in Sec.\ref{subsec:Cases Study}, we follow the original paper\cite{libai_corridor} and deploy OCP-STC using AMPL\cite{ampl}.

\begin{figure}[t]
    \centering
    \includegraphics[width=1.0\columnwidth]{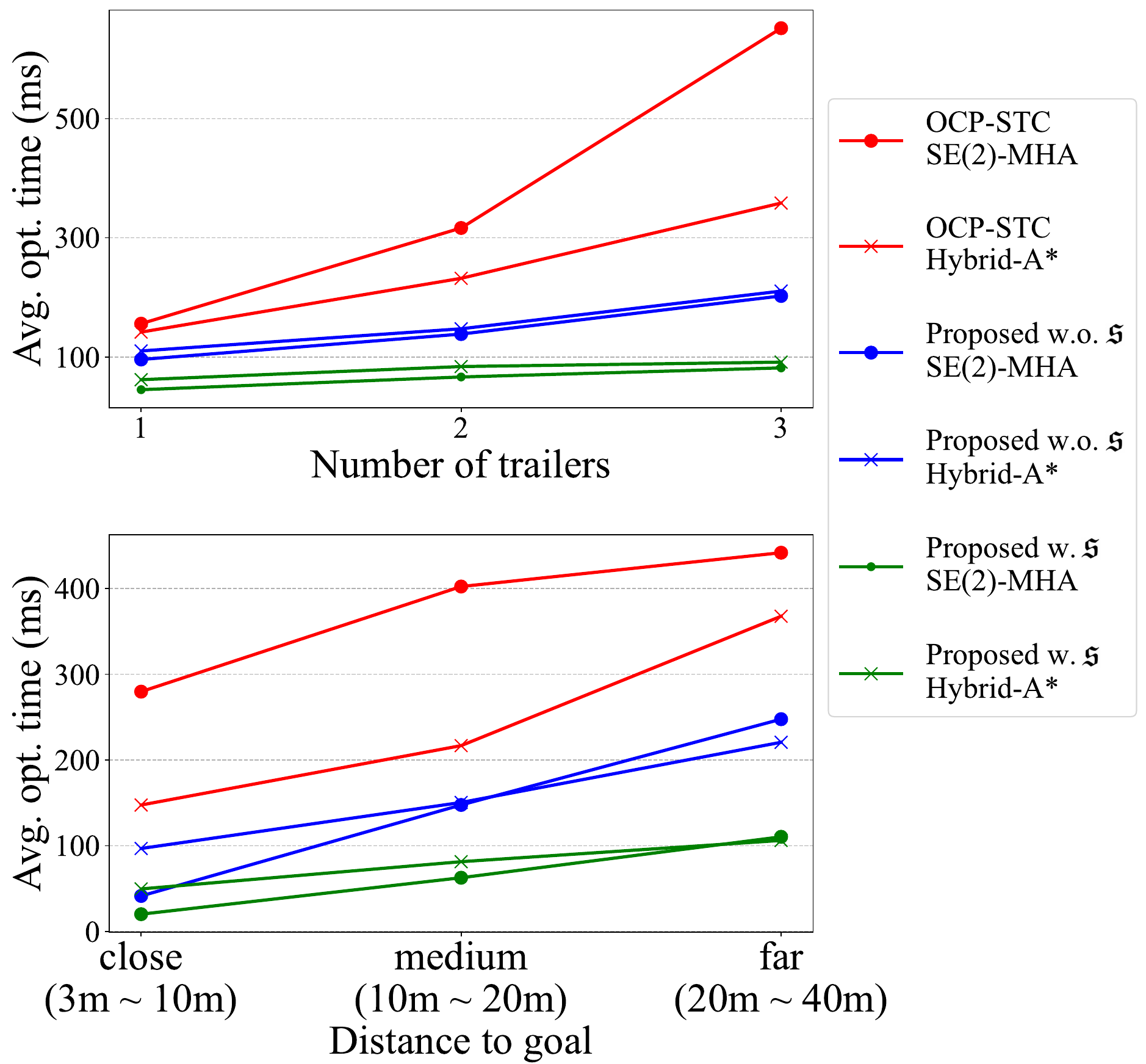}
	\caption{Line graphs of average optimization time as a function of problem dimensions in the environment with a medium density of obstacles.}\label{fig:opt_curv}
\end{figure}

Two initial value acquisition methods and three optimization methods are tested in the above-mentioned environments. The initial value acquisition methods are the proposed SE(2)-MHA and Hybrid-A*, where the latter is the path search method proposed in the work\cite{libai_pcoc}, which does exploration over the entire $\mathbb{R}\times SO(2)^\text{N+1}$ state space. The optimization methods include OCP-STC\cite{libai_corridor}, Proposed w.o. $\mathfrak{s}$ and Proposed w. $\mathfrak{s}$, where w.o. $\mathfrak{s}$ or w. $\mathfrak{s}$ is with or without using the \textit{slackened arc length} as optimization variables. We compare these method in different environments with different density of obstacles when number of trailers $\text{N}$ is 1, 2, or 3. Besides, tasks with different distance from the center of the rear wheels of the tractor to the center point of the target area ($3m\sim10m$, $10m\sim20m$, $20m\sim40m$) are conducted separately. More than one hundred comparison tests are performed in each case.

Fig.~\ref{fig:opt_curv} illustrates how the trajectory optimization time varies with the problem dimensions. Regardless of which method is used to obtain the initial values, Proposed with \textit{slackened arc length} $\mathfrak{s}$ consumes the least amount of time in optimization. As the number of trailers increases and the target region becomes farther away, this method also shows a slower increase in computational time consumption. This is due to the fact that for the similar problem dimension, compared to OCP-STC\cite{libai_corridor}, we utilize polynomial trajectories and differential flatness to reduces the dimensionality of the optimization variables. Meanwhile, despite the fact that we introduce \textit{slackened arc length} and increase optimization variables, the constructed trajectory optimization problem has no numerical problems caused by singularities, which in turn makes the optimization easier to converge to the local optimum and thus shortens the optimization time. Without \textit{slackened arc length}, the effects from singularities are more and more exposed as the problem becomes more complex and more constraints are imposed. Thus the optimization efficiency decreases drastically. In addition, it can also be seen from Fig.~\ref{fig:opt_curv} that the dependency of OCP-STC\cite{libai_corridor} on the initial value is strong. The collision-free path provided by Hybrid-A* makes it possible to reduce the optimization time significantly.

\begin{figure}[t]	
    \centering
    \includegraphics[width=1.0\columnwidth]{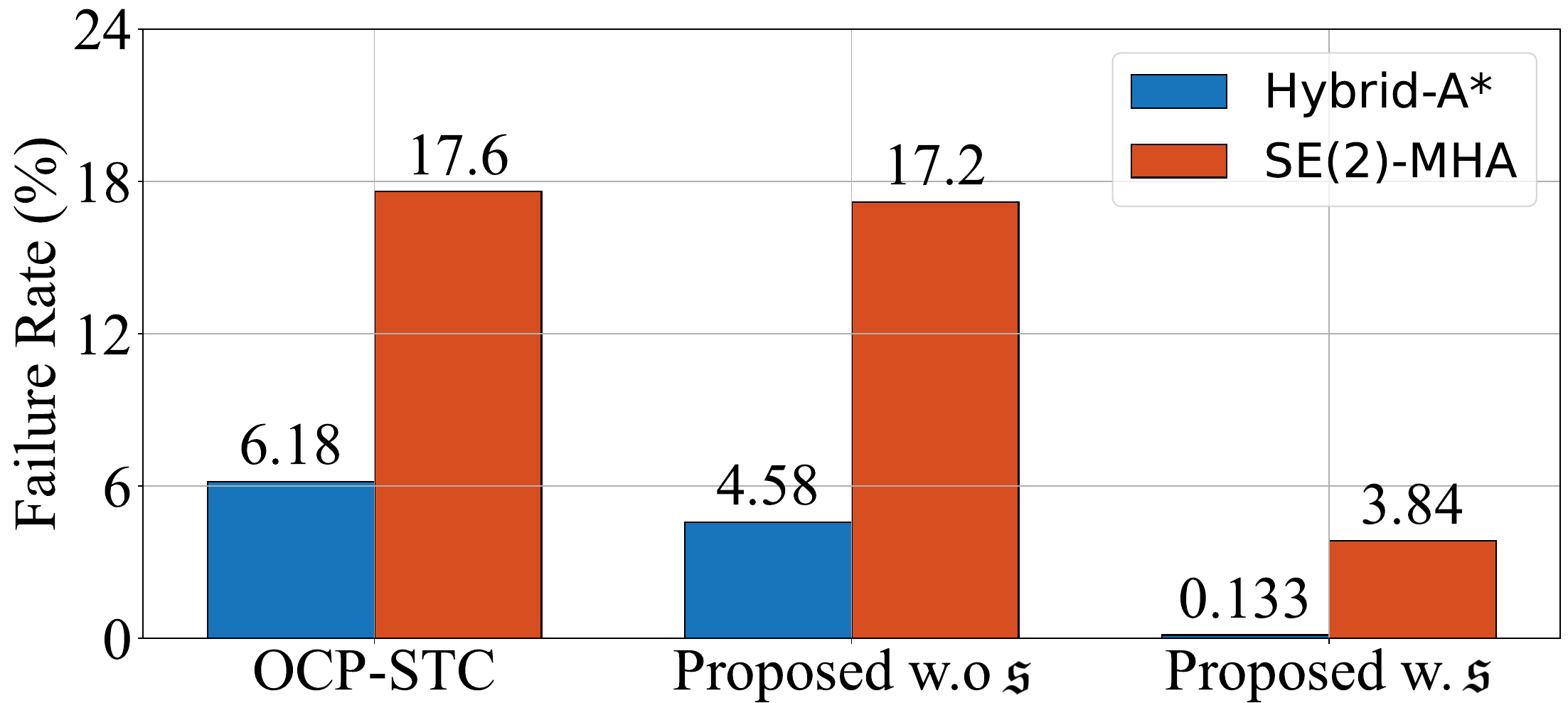}
	\caption{Failure rate in the environment with a medium density of obstacles.}\label{fig:succ}
\end{figure}

\begin{table*}
\small
\centering
\renewcommand\arraystretch{1.2}
\caption{\label{tab:bk} Benchmark Comparison in Different Cases }
\begin{tabular}{c|ccccccccccc}
\hline
\multirow{2}{*}{$\text N$} &
  \multicolumn{2}{c}{Environments} &
  \multicolumn{3}{c}{\begin{tabular}[c]{@{}c@{}}Low-Complexity\\ (10, 10, 10)\end{tabular}} &
  \multicolumn{3}{c}{\begin{tabular}[c]{@{}c@{}}Medium-Complexity\\ (20, 20, 20)\end{tabular}} &
  \multicolumn{3}{c}{\begin{tabular}[c]{@{}c@{}}High-Complexity\\ (40, 40, 40)\end{tabular}} \\ \cline{2-12} 
 &
  \multicolumn{2}{c}{Method} &
  $l_\text{traj}$ (m) &
  $t_\text{d}$ (s) &
  $\kappa$(m$^{-1}$) &
  $l_\text{traj}$ (m) &
  $t_\text{d}$ (s) &
  $\kappa$(m$^{-1}$) &
  $l_\text{traj}$ (m) &
  $t_\text{d}$ (s) &
  $\kappa$(m$^{-1}$) \\ \hline
\multirow{6}{*}{1} &
  \multicolumn{1}{c|}{\multirow{3}{*}{\begin{tabular}[c]{@{}c@{}}Hybrid-A*\\ (89.57\%)\\ (926.2ms)\end{tabular}}} &
  \multicolumn{1}{c|}{OCP-STC} &
  17.80 &
  11.68 &
  0.3267 &
  17.98 &
  12.83 &
  0.3381 &
  18.68 &
  15.08 &
  0.3593 \\
 &
  \multicolumn{1}{c|}{} &
  \multicolumn{1}{c|}{Proposed w.o. $\mathfrak{s}$} &
  17.46 &
  10.58 &
  0.3829 &
  17.38 &
  10.55 &
  0.4198 &
  17.55 &
  10.60 &
  0.4163 \\
 &
  \multicolumn{1}{c|}{} &
  \multicolumn{1}{c|}{Proposed w. $\mathfrak{s}$} &
  \textbf{17.27} &
  \textbf{9.959} &
  \textbf{0.2489} &
  \textbf{17.23} &
  \textbf{9.940} &
  \textbf{0.2595} &
  \textbf{17.39} &
  \textbf{10.00} &
  \textbf{0.2603} \\ \cline{2-3}
 &
  \multicolumn{1}{c|}{\multirow{3}{*}{\begin{tabular}[c]{@{}c@{}}SE(2)-MHA\\ (98.49\%)\\ (115.7ms)\end{tabular}}} &
  \multicolumn{1}{c|}{OCP-STC} &
  17.69 &
  11.24 &
  0.3389 &
  18.44 &
  13.75 &
  0.3458 &
  19.32 &
  20.04 &
  0.5986 \\
 &
  \multicolumn{1}{c|}{} &
  \multicolumn{1}{c|}{Proposed w.o. $\mathfrak{s}$} &
  17.32 &
  10.61 &
  0.4935 &
  17.29 &
  10.56 &
  0.5095 &
  17.84 &
  11.11 &
  0.5504 \\
 &
  \multicolumn{1}{c|}{} &
  \multicolumn{1}{c|}{Proposed w. $\mathfrak{s}$} &
  {\ul 17.30} &
  {\ul 10.05} &
  {\ul 0.2469} &
  \textbf{17.23} &
  {\ul 9.956} &
  {\ul 0.2474} &
  {\ul 17.72} &
  {\ul 10.24} &
  {\ul 0.2733} \\ \hline
\multirow{6}{*}{2} &
  \multicolumn{1}{c|}{\multirow{3}{*}{\begin{tabular}[c]{@{}c@{}}Hybrid-A*\\ (84.80\%)\\ (904.6ms)\end{tabular}}} &
  \multicolumn{1}{c|}{OCP-STC} &
  18.14 &
  11.41 &
  0.2687 &
  18.76 &
  13.65 &
  0.2955 &
  18.87 &
  17.79 &
  0.2948 \\
 &
  \multicolumn{1}{c|}{} &
  \multicolumn{1}{c|}{Proposed w.o. $\mathfrak{s}$} &
  17.88 &
  10.80 &
  0.3629 &
  18.11 &
  11.09 &
  0.3609 &
  {\ul 17.75} &
  10.81 &
  0.3735 \\
 &
  \multicolumn{1}{c|}{} &
  \multicolumn{1}{c|}{Proposed w. $\mathfrak{s}$} &
  \textbf{17.58} &
  \textbf{10.10} &
  \textbf{0.2353} &
  \textbf{17.84} &
  \textbf{10.23} &
  \textbf{0.2395} &
  \textbf{17.53} &
  \textbf{10.07} &
  \textbf{0.2416} \\ \cline{2-3}
 &
  \multicolumn{1}{c|}{\multirow{3}{*}{\begin{tabular}[c]{@{}c@{}}SE(2)-MHA\\ (98.53\%)\\ (132.1ms)\end{tabular}}} &
  \multicolumn{1}{c|}{OCP-STC} &
  18.57 &
  12.18 &
  0.3017 &
  19.92 &
  18.11 &
  0.3171 &
  20.49 &
  29.39 &
  0.3401 \\
 &
  \multicolumn{1}{c|}{} &
  \multicolumn{1}{c|}{Proposed w.o. $\mathfrak{s}$} &
  18.02 &
  11.27 &
  0.5010 &
  18.29 &
  11.19 &
  0.4965 &
  18.75 &
  11.44 &
  0.4731 \\
 &
  \multicolumn{1}{c|}{} &
  \multicolumn{1}{c|}{Proposed w. $\mathfrak{s}$} &
  {\ul 17.69} &
  {\ul 10.17} &
  {\ul 0.2470} &
  {\ul 18.04} &
  {\ul 10.34} &
  {\ul 0.2410} &
  18.49 &
  {\ul 10.56} &
  {\ul 0.2535} \\ \hline
\multirow{6}{*}{3} &
  \multicolumn{1}{c|}{\multirow{3}{*}{\begin{tabular}[c]{@{}c@{}}Hybrid-A*\\ (75.05\%)\\ (887.7ms)\end{tabular}}} &
  \multicolumn{1}{c|}{OCP-STC} &
  19.30 &
  12.72 &
  0.2799 &
  19.10 &
  13.90 &
  0.2710 &
  19.44 &
  18.25 &
  0.3619 \\
 &
  \multicolumn{1}{c|}{} &
  \multicolumn{1}{c|}{Proposed w.o. $\mathfrak{s}$} &
  18.75 &
  11.35 &
  0.3489 &
  {\ul 18.21} &
  10.99 &
  0.3475 &
  {\ul 18.24} &
  11.11 &
  0.2844 \\
 &
  \multicolumn{1}{c|}{} &
  \multicolumn{1}{c|}{Proposed w. $\mathfrak{s}$} &
  18.46 &
  {\ul 10.55} &
  \textbf{0.2231} &
  \textbf{17.94} &
  \textbf{10.28} &
  \textbf{0.2156} &
  \textbf{17.97} &
  \textbf{10.29} &
  \textbf{0.2274} \\ \cline{2-3}
 &
  \multicolumn{1}{c|}{\multirow{3}{*}{\begin{tabular}[c]{@{}c@{}}SE(2)-MHA\\ (98.50\%)\\ (109.8ms)\end{tabular}}} &
  \multicolumn{1}{c|}{OCP-STC} &
  19.79 &
  15.39 &
  0.3330 &
  20.36 &
  22.64 &
  0.3136 &
  20.87 &
  30.71 &
  0.4271 \\
 &
  \multicolumn{1}{c|}{} &
  \multicolumn{1}{c|}{Proposed w.o. $\mathfrak{s}$} &
  {\ul 18.26} &
  11.41 &
  0.5202 &
  18.48 &
  11.47 &
  0.4662 &
  18.92 &
  11.78 &
  0.4899 \\
 &
  \multicolumn{1}{c|}{} &
  \multicolumn{1}{c|}{Proposed w. $\mathfrak{s}$} &
  \textbf{17.96} &
  \textbf{10.33} &
  {\ul 0.2376} &
  {\ul 18.21} &
  {\ul 10.44} &
  {\ul 0.2317} &
  18.57 &
  {\ul 10.61} &
  {\ul 0.2540} \\ \hline
\end{tabular}
\end{table*}

Fig.~\ref{fig:succ} shows the effect of different initial value selection methods on the success rate of the three different methods when the obstacle density is medium. Where OCP-STC\cite{libai_corridor} is considered to fail after the optimization consumes more than 30 seconds. Proposed w.o./w. $\mathfrak{s}$ is considered to fail when the iterations of PHR-ALM is greater than 50. Since Hybrid-A* searches for paths in the full state and can provide collision-free initial values, the optimizers all have a higher success rate compared to using SE(2)-MHA. However, the need to generate corridors for all vehicles based on the initial values may lead to problems infeasible in some narrower environments, which lead to higher failure rates when optimization method is OCP-STC\cite{libai_corridor}. Without $\mathfrak{s}$, singularities lead to poor initial values of the optimization problem, resulting in the optimizer may not converge in a finite number of steps or there is a probability of optimizing to a local minima that cannot satisfy the constraints. Thus there is also a higher failure rate when using Proposed w.o. $\mathfrak{s}$. In the case of using SE(2)-MHA, it may not be possible to generate corridors by initial value as in Fig.~\ref{fig:park}, leading to a significantly higher failure rate of using OCP-STC\cite{libai_corridor}. The poorer initial values also amplify the problem of singularities to Proposed w.o. $\mathfrak{s}$ even more, leading to higher failure rates. SE(2)-MHA also leads to more failures in Proposed w. $\mathfrak{s}$, which is mainly attributed to the incompleteness of SE(2)-MHA. It sacrifices the feature of collision-free path for sake of efficiency and thus has a higher probability of generating initial values that make it impossible for the optimizer to optimize to a feasible local minima.

More results on efficiency and quality are summarized in Table \ref{tab:bk}. Where the tuple $(x,y,z)$ under environment complexity denotes the number of triangular, quadrilateral, and pentagonal random obstacles respectively. $l_\text{traj}$ denotes the mean length of the tractor trajectory, $t_\text{d}$ denotes the mean time duration of the optimized trajectory, and $\kappa$ denotes the mean curvature of the tractor, reflecting the smoothness of the trajectory. We set the front-end path search of greater than 5 seconds to count as failure. From the table, it can be seen that the success rate and computation time of SE(2)-MHA do not change much as the number of trailers becomes more. Its success rates are all greater than 98\% and time consumption is all below 140 ms. This is attributed to the fact that it only searches in the $SE(2)$ space while utilizing the numerical integration approximation to obtain the trailers' trajectory. On the contrary, since Hybrid-A* has to do the search and collision detection in the complete state space, the success rate decreases with the increase of the number of trailers under the time limit of 5 seconds. It far exceeds SE(2)-MHA in terms of average time consumption, which is at least seven times more than the latter.

In Table~\ref{tab:bk}, Hybrid-A* + Proposed w. $\mathfrak{s}$ performs best in most cases. Hybrid-A* searches the space more thoroughly, the initial values obtained tend to be better than SE(2)-MHA. In most cases, Proposed w. $\mathfrak{s}$ initialized with SE(2)-MHA performs second best and has a smaller gap with Hybrid-A* + Proposed w. $\mathfrak{s}$. Although the trajectory length and trajectory duration of Proposed w.o. $\mathfrak{s}$ are slightly better than OCP-STC\cite{libai_corridor} in most cases, the average curvature of the tractor trajectories is the largest in many cases due to the inability to handle the singularity. Since the strategy of generating corridors for each vehicle may lead to a smaller feasible space containing less optimal solutions, OCP-STC\cite{libai_corridor} does not perform as well as the proposed methods in terms of trajectory length and trajectory duration.

\begin{figure}[t]
    \centering
    \includegraphics[width=1.0\columnwidth]{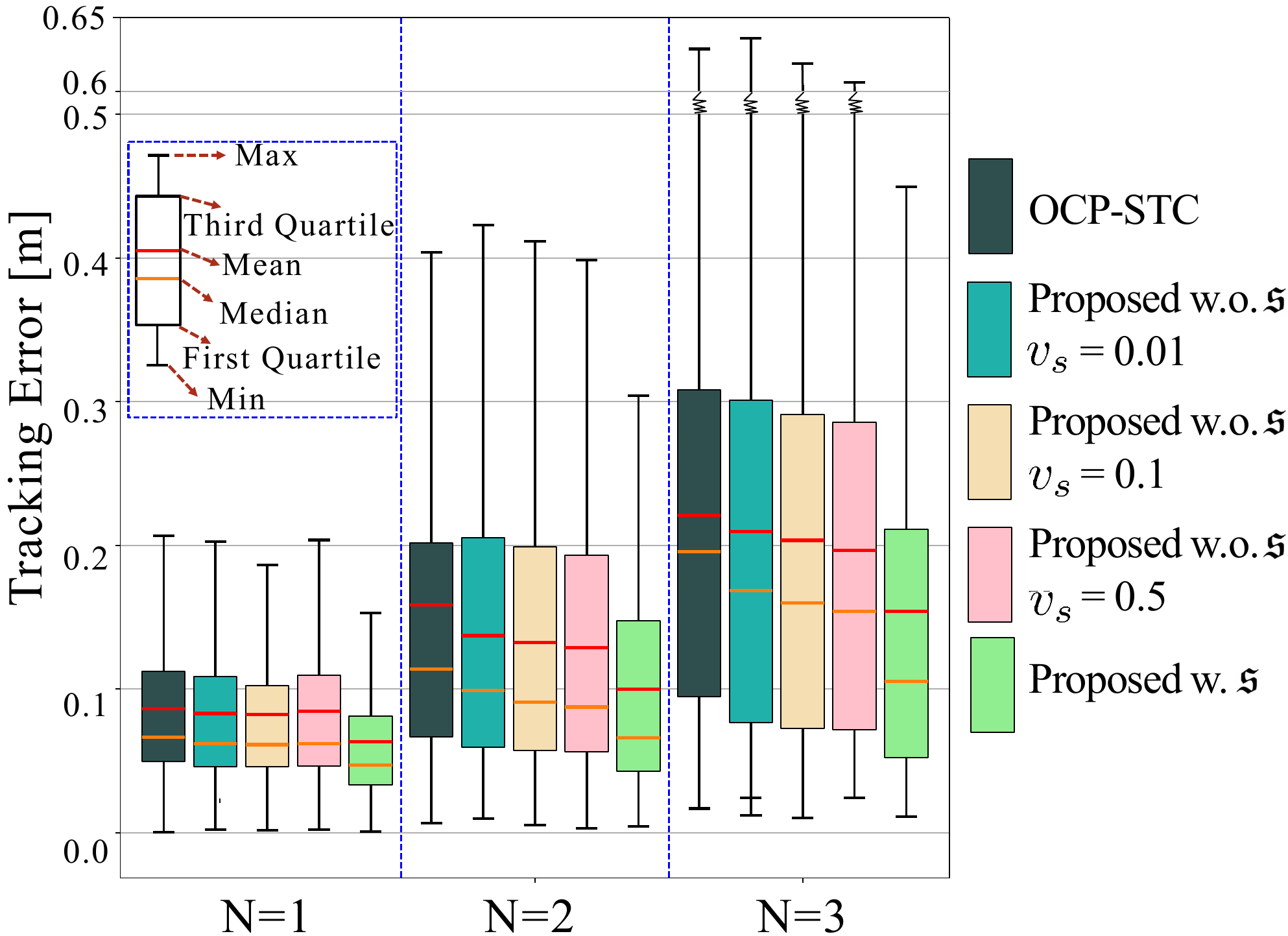}
    \caption{The statistical analysis of tracking errors for each trajectory during their execution.}\label{fig:bk_track}
\end{figure}

To further evaluate the execution performance of trajectories, using Gazbeo\cite{gazebo}, we model the robot used in the real-world experiments into the physical simulation shown in Fig.~\ref{fig:gazebo}, recording more than one hundred trajectories for each of the robots with different numbers of trailers to perform the tracking error test as in the ablation experiment. The results are shown in Fig.~\ref{fig:bk_track}. We can see that the trajectory tracking error decreases by almost $20\%$ with the proposed method, comparing without using the slackened arc length $\mathfrak{s}$. In addition, compared to OCP-STC\cite{libai_corridor}, our method also achieves lower tracking errors. We attribute this to the high-order continuous polynomial trajectory representation. Compared to methods employing discrete motion processes, our approach essentially guarantees the continuity of the state and finite-dimensional derivatives, making our trajectories closer to real physical motion processes, and thus easier to execute.

\begin{figure}[t]
    \centering
    \includegraphics[width=1.0\columnwidth]{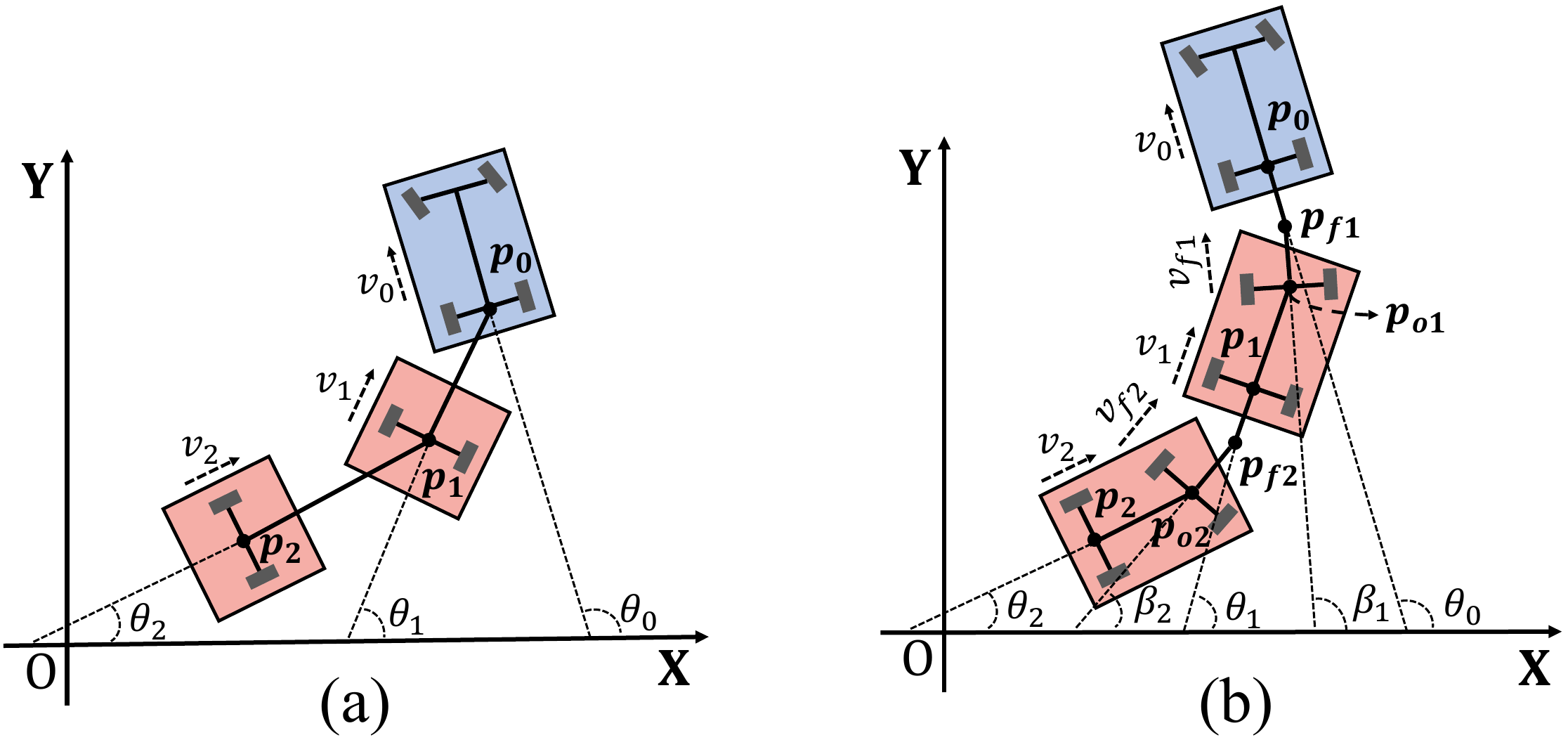}
    \caption{Different tractor-trailer models studied in our work (a) and the work~\cite{system_opt} (b).}\label{fig:new_model}
\end{figure}

\section{Discussion}
\subsection{Generalizability}
In this work, we focus on tractor-trailer robots with trailers having only two passive wheels, as shown in Fig.~\ref{fig:new_model}(a). However, there are also tractor-trailer robots with trailers having four passive wheels in the real world, as shown in Fig.~\ref{fig:new_model}(b), which had been studied in the work~\cite{system_opt}. In this setup, each trailer has four passive wheels, two axles, an on-axle hitch (points $\boldsymbol{p}_{o1}$ and $\boldsymbol{p}_{o2}$) and an off-axle hitch (points $\boldsymbol{p}_{f1}$ and $\boldsymbol{p}_{f2}$). This trailer model is more elaborate with the advantage of reducing the forces on the linkage, while making the motion of the trailer smoother and less likely to result in jackknife\cite{jackknife}. Despite such a more complex trailer model, with two more degrees of freedom compared to the model we studied, our method can still be quickly transferred on it, with only simple modifications to the optimization variables and constraints. Taking the example of a tractor with two identical trailers, the kinematic model for a robot similar to the one in the work~\cite{system_opt} is:

\begin{align}
\dot x_0(t)=&v_0(t)\cos\theta_0(t),\\
\dot y_0(t)=&v_0(t)\sin\theta_0(t),\\
\dot v_0(t)=&a(t),\\
\dot \theta_0(t)=&v_0(t)\frac{\tan\delta(t)}{L_0},
\end{align}
\begin{align}
\dot \beta_1(t)=&\frac{v_0(t)\sin(\theta_0(t)-\beta_1(t))}{L_h}\nonumber\\
&-\frac{L_t\dot\theta_0(t)\cos(\theta_0(t)-\beta_1(t))}{L_h},\label{eq:newbegin}\\
\dot \theta_1(t)=&\frac{v_{f1}(t)\sin(\beta_1(t)-\theta_1(t))}{L_w},\\
\dot \beta_2(t)=&\frac{v_1(t)\sin(\theta_1(t)-\beta_2(t))}{L_h}\nonumber\\
&-\frac{L_t\dot\theta_1(t)\cos(\theta_1(t)-\beta_2(t))}{L_h},\\
\dot \theta_2(t)=&\frac{v_{f2}(t)\sin(\beta_2(t)-\theta_2(t))}{L_w},\label{eq:newend}
\end{align}
where $v_0,a,\delta,L_0$ are the velocity, longitudinal acceleration, steering angle, and wheelbase length of the tractor, respectively. $L_t$ is the distance from the rear center of the previous car to the off-axle hitch. $L_h$ is the distance from the off-axle hitch to the on-axle hitch in a trailer. $L_w$ is the wheelbase length of a trailer. $v_{f1}=v_0\cos(\theta_0-\beta_1)+L_t\dot\theta_0\sin(\theta_0-\beta_1),v_1=v_{f1}\cos(\beta_1-\theta_1),v_{f1}=v_1\cos(\theta_1-\beta_2)+L_t\dot\theta_1\sin(\theta_1-\beta_2)$. 

\begin{figure}[t]
    \centering
    \includegraphics[width=1.0\columnwidth]{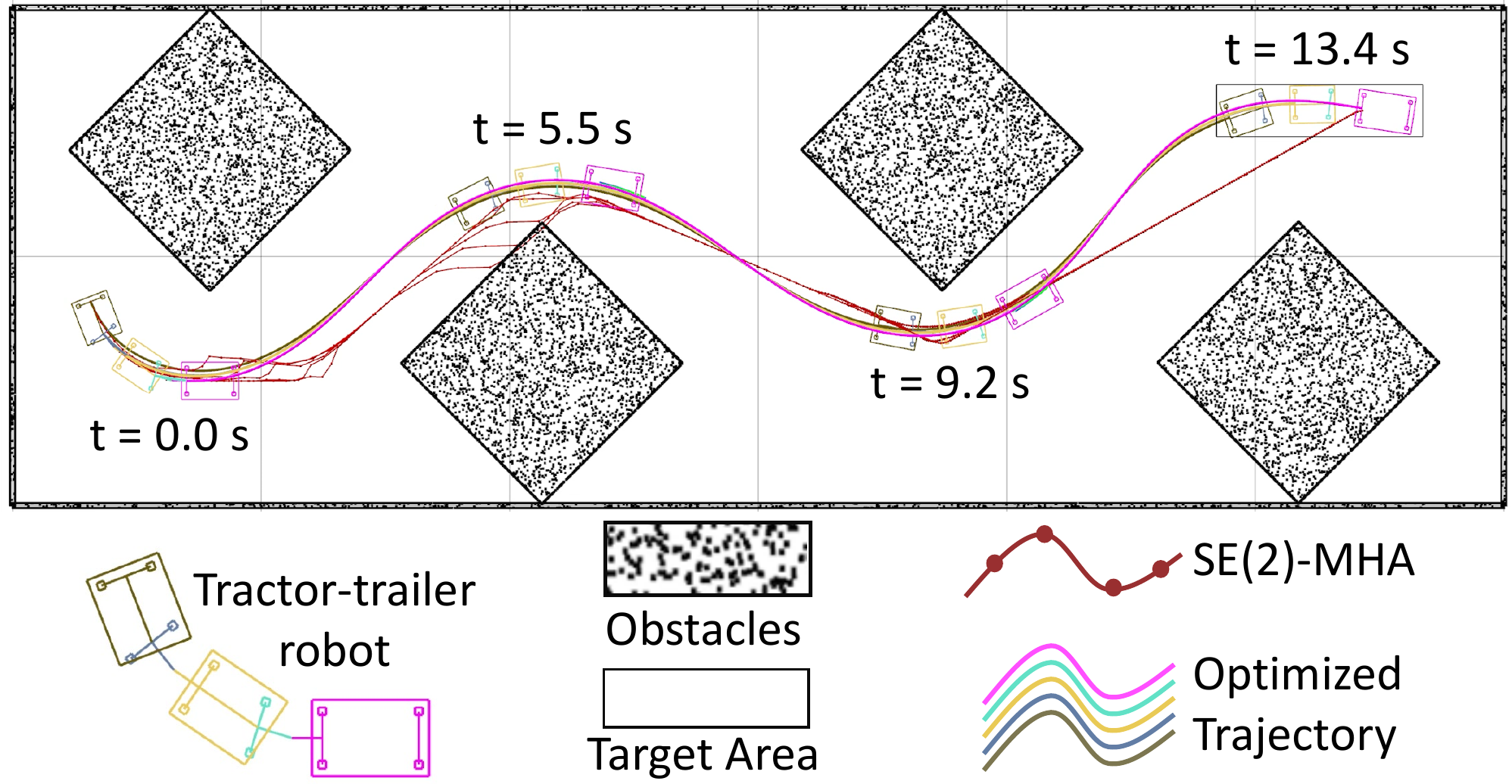}
    \caption{Our pipeline can generalize to the more complex model studied in work~\cite{system_opt}.}\label{fig:new_model_test}
\end{figure}

Comparing the model we studied, by simply introducing just two new trajectories for $\beta_1$ and $\beta_2$, and modifying the equation constraints for kinematic constraints (\ref{con:eqc}) to those for kinematic constraints (\ref{eq:newbegin}) $\sim$ (\ref{eq:newend}) in the trajectory optimization problem, we are able to perform motion planning for this kinds of robot. As shown in Fig.~\ref{fig:new_model_test}, we performed a simulation experiment in the ROS environment. In this scenario, the robot is asked to traverse an area containing obstacles to reach the target area. The code targeting this complex trailer model is also open-sourced in a branch\footnote{\url{https://github.com/Tracailer/Tracailer/tree/general_model}} in our Github repository as a reference for the community.

\begin{figure*}[t]
    \centering
    \includegraphics[width=1.0\linewidth]{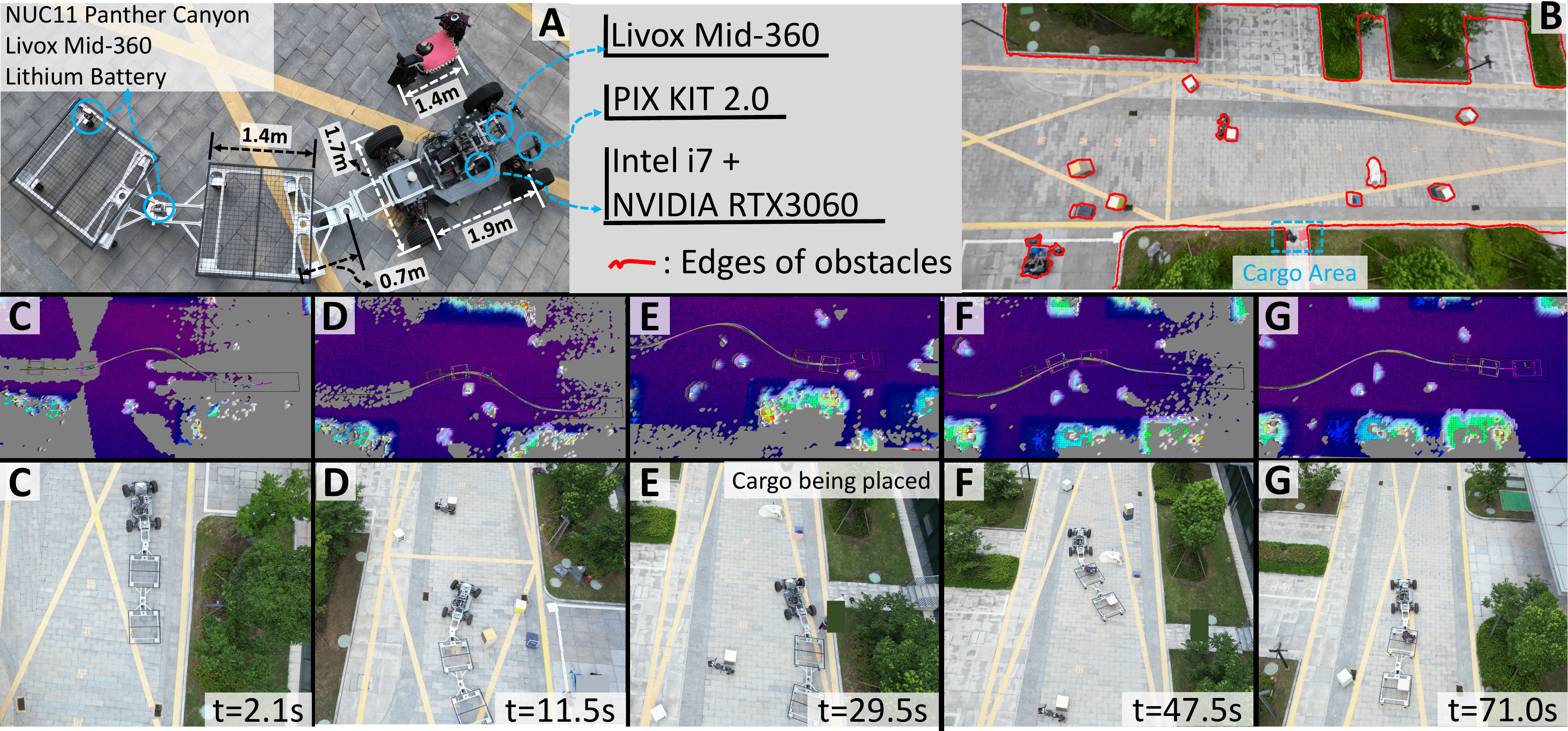}
    \caption{Real-world experiments for the more complex model studied in work~\cite{system_opt}. Figure A shows the configuration and size of the robot. Figure B shows the top view of the experimental scenario. Figure C $\sim$ G show the third-person viewpoints or corresponding views from Rviz tacken at different moments.}\label{fig:bigcar}
\end{figure*}

To further demonstrate the generalizability of our algorithm, we conduct real-world experiments on a larger-scale robot. we selected the \textit{PIX KIT 2.0} chassis as the tractor, designing trailers similar to the one studied in work~\cite{system_opt}, as shown in Fig.~\ref{fig:bigcar} A. In this experiment, the kinematic constraints of the tractor are set to the maximum longitude velocity $v_\text{mlon}=1.0m/s$, the maximum longitude acceleration $a_\text{mlon}=0.8m/s^2$, the maximum latitude acceleration $a_\text{mlat}=1.2m/s^2$, and maximum curvature $\kappa_\text{max}=0.28m^{-1}$. To have smoother control commands, we also limit the maximum steering velocity $\dot\delta_\text{max}=0.6rad/s$ of the tractor in trajectory optimization. 

Since the control interfaces of the chassis are throttle, brake and steering angle, we modify the control output of the trajectory tracking controller to be acceleration and steering angle, interpolating the throttle-brake calibration table corresponding to speed and acceleration to obtain the final throttle or brake commands. As for the localization module, we place a LiDAR and run LIO on each trailer separately, aligning the world frame with the LIO running on the tractor. Knowing the absolute position of neighboring vehicles, we can use the cosine theorem to calculate the yaw angle $\beta_1,\beta_2$ of the wheel axle corresponding to the on-axle hitch. 

We place e-bike, rockery, boxes as obstacles on a path of an industrial park and set up a cargo area, as shown in Fig.~\ref{fig:bigcar} B. The robot is requested to avoid obstacles to reach near the cargo area. After the staff load the cargo, it is required to continue traversing the obstacle area, as detailed in Fig.~\ref{fig:bigcar} C $\sim$ G. A more complete process can be found in the attached multimedia.

\begin{figure*}[t]
    \centering
    \includegraphics[width=1.0\linewidth]{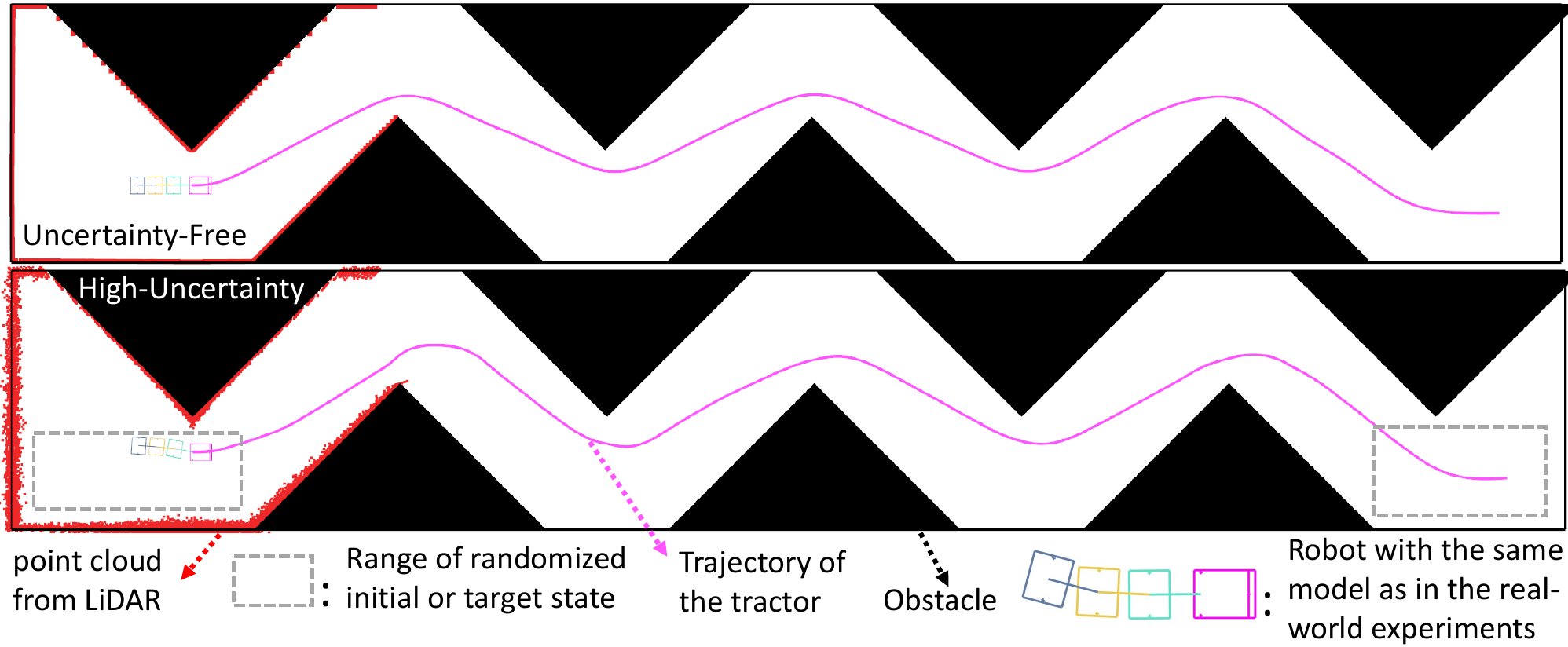}
    \caption{Motion trajectory visualization under cases without and with high uncertainty.}\label{fig:noise}
\end{figure*}

\begin{table*}[t]
	\small
	\centering
	\renewcommand\arraystretch{1.2}
	\caption{\label{tab:noise} Statistics of Trajectory Dynamics Under Various Uncertainties.}
        \begin{tabular}{|c|cccc|cccccc|}
    \hline
    \multirow{2}{*}{Uncertainty} &
      \multirow{2}{*}{\begin{tabular}[c]{@{}c@{}}$\sigma_d$\\ (m)\end{tabular}} &
      \multirow{2}{*}{\begin{tabular}[c]{@{}c@{}}$\sigma_\theta$\\ (Deg)\end{tabular}} &
      \multirow{2}{*}{\begin{tabular}[c]{@{}c@{}}Delay\\ (ms)\end{tabular}} &
      \multirow{2}{*}{Model} &
      \multirow{2}{*}{M.R} &
      \multirow{2}{*}{\begin{tabular}[c]{@{}c@{}}S.R\\ ($\%$)\end{tabular}} &
      \multirow{2}{*}{\begin{tabular}[c]{@{}c@{}}M.TE\\ ($cm$)\end{tabular}} &
      \multirow{2}{*}{\begin{tabular}[c]{@{}c@{}}M.TL\\ ($m$)\end{tabular}} &
      \multirow{2}{*}{\begin{tabular}[c]{@{}c@{}}M.V\\ ($m/s$)\end{tabular}} &
      \multirow{2}{*}{\begin{tabular}[c]{@{}c@{}}M.SR\\ (rad$/s$)\end{tabular}} \\
           &      &   &      &                     &     &     &      &       &      &       \\ \hline
    Free   & 0    & 0 & 0    & {[}0, 0{]}          & 33  & 100 & 1.67 & 55.80 & 0.98 & 0.030 \\ \hline
    Low    & 0.05 & 1 & 0.02 & {[}-0.05, 0.15{]}   & 34  & 100 & 2.73 & 55.94 & 0.93 & 0.033 \\ \hline
    Medium & 0.1  & 2 & 0.05 & {[}-0.075, 0.225{]} & 43  & 100 & 5.07 & 56.18 & 0.87 & 0.042 \\ \hline
    High   & 0.15 & 4 & 0.07 & {[}-0.1, 0.3{]}     & 248 & 84  & 9.59 & 58.45 & 0.45 & 0.060 \\ \hline
    \end{tabular}
\end{table*}

\subsection{Tests under various uncertainty and noise}
To further evaluate the impact of various uncertainties and noises on the proposed planner, we conduct simulation experiments in a $60m*10m$ unknown environment, as shown in Fig.~\ref{fig:noise}. The tractor-trailer robot is configured with the same parameters as used in our real-world experiments and is equipped with an omnidirectional radar capable of sensing objects up to $10m$ long. After the planner outputs a trajectory, we use a MPC-based controller\cite{mpc} to track it.

In the real world, the sources of uncertainty and noise are mainly caused by the perception and control modules. The uncertainty in the former mainly comes from the uncertainty in the perception of obstacles in the environment, and the uncertainty in the estimation of robot's state. While the uncertainty of the control module mainly originates from the uncertainty of the robot model. Thus, we add the noises from the following three aspects:
\begin{itemize}
\item [1)] \textbf{Uncertainty in environment perception:} In our experiments, the robot's perception of the environment comes from the LiDAR. So we add noise to the distance scanned by the LiDAR: $\hat{d}_i=d_i+n_{di},i=1,2,...,N_l$, where $d_i$ is the ground truth of the distance, $N_l$ is the number of distances scanned by the LiDAR. $n_{di}\sim\mathcal{N}(0,\sigma_d^2)$, $\sigma_d>0$ is the standard deviation of the normal distribution.
\item [2)] \textbf{Uncertainty in robot state perception:} Due to the deformability of the tractor-trailer robots, we impose noise on the yaw angle estimation of the trailers in the
\end{itemize}
\begin{itemize}
\item [] robot: $\hat{\theta}_i-\hat{\theta}_{i+1}=\theta_i-\theta_{i+1}+n_{\theta i},i=0,1,...,N_t$, where $\theta_i$ is the ground truth of the yaw angle, $\theta_0=\hat{\theta}_0$ is the yaw angle of the tractor, $N_t+1=3$ is the number of trailers, $n_{\theta i}\sim\mathcal{N}(0,\sigma_\theta^2)$.
\item [3)] \textbf{Uncertainty in robot model:} Referring to the experiments on model uncertainty in the work~\cite{hzctits}, the kinematic model with noise $\boldsymbol{n}_m$ used in this experiment is as follows:
\begin{align}
\dot{\boldsymbol{s}}&=f(\boldsymbol{s},\boldsymbol{u},\boldsymbol{n}_m),\\
\boldsymbol{s}&=[x_0,y_0,v_0,\theta_0,\theta_1,\theta_2,\theta_3]^\text{T},\\
\boldsymbol{u}&=[a,\delta]^\text{T},
\end{align}
where $[x_0,y_0]$ is the position of the rear center of the tractor, $v_0$ is the longitude velocity, $a$ is the longitude acceleration, $\delta$ is the steer angle, $\theta_0$ is yaw angle of the tractor,  $\theta_i,i=\{1,2,3\}$ are yaw angles of the trailers, respectively. Here, the state transfer equation $f$ with noise is defined as:
\begin{align}
f(\boldsymbol{s},\boldsymbol{u},\boldsymbol{n}_m)=&q(\boldsymbol{s},\boldsymbol{u})+N(\boldsymbol{s},\boldsymbol{u},\boldsymbol{n}_m),\\
q(\boldsymbol{s},\boldsymbol{u})=&[v_0\cos\theta_0,v_0\sin\theta_0,a,v_0\tan\delta/L_w,\nonumber\\
&v_0\sin(\theta_0-\theta_1)/L_t,\nonumber\\
&v_1\sin(\theta_1-\theta_2)/L_t,\nonumber\\
&v_2\sin(\theta_2-\theta_3)/L_t]^\text{T},
\end{align}
where $v_1=v_0\cos(\theta_0-\theta_1),v_2=v_1\cos(\theta_1-\theta_2)$. $L_w$ is the wheelbase length of the tractor, $L_t$ is rigid link length of the trailer. $q$ is the theoretical state transfer, and $N = \boldsymbol{n}_mq$ is the uncertainty term defined as proportional to the change in motion. $\boldsymbol{n}_m$ is a diagonal matrix representing the intensity of random noise, where each element satisfies a uniform distribution on
[$h_1, h_2$]. Also, delays are introduced between the controller and actuator to better simulate real-world scenarios.
\end{itemize}

In this experiment, replanning is triggered if the original trajectory collides with a newly detected obstacle, or the tracking error exceeds $0.1m$. Robots hitting obstacles during movement are considered as failed attempts. The experiment cases are classified into four groups based on the levels of noise and delay: free uncertainty, low uncertainty, medium uncertainty, and high uncertainty. We refer to the datasheets and specifications of common LiDARs, such as \textit{Livox MID-360}, Ouster\cite{ouster} to select $\sigma_d=5cm$ for the low uncertainty case. The specifications of the absolute encoders from \textit{OMRON} and \textit{HEIN LANZ} are referenced to set $\sigma_\theta=1^\circ$ for the low uncertainty case. Finally, we set the delays and noise for the different cases by referring to the experiments on model uncertainty in the work~\cite{hzctits}.

One hundred tests are conducted for each case. Moreover, success rate (S.R), mean number of replans (M.R), mean tracking error (M.TE), mean trajectory length (M.TL), mean velocity (M.V), and mean steering angle rate (M.SR) are calculated. The quantitative results are summarized in Table~\ref{tab:noise}. Despite the presence of uncertainty, the robot still exhibits some resilience due to its high-frequency trajectory tracking controller and replanning capability, allowing it to navigate safely in unknown environments with $84\%$ success rate even under high uncertainty. The noise intensity in this test environment actually exceeds that of the vast majority in the real world. Such rigorous simulations validate the robustness of our pipeline.

The results in the Table~\ref{tab:noise} further indicate that high uncertainty leads to more frequent excessive tracking errors, which in turn results in a dramatic increase in the number of replans. On the other hand, replanning techniques allow the robot to eliminate accumulated tracking errors, thus enhancing its robustness against uncertainty. However, the increase in uncertainty still has some other impact. As shown by the increase in mean tracking error and steering rate, it becomes more difficult to track the trajectory. Second, smaller mean velocities and longer trajectories imply reduced optimality. In the future, we will incorporate the potential uncertainties in the controllers and the proposed planner, with the expectation of further increasing the practical applicability and the resilience of the system to uncertainty.

\section{Conclusion}
\label{sec:Conclusion}
In this paper, we propose a new trajectory representation for tractor-trailer robot. Based on it, we formulate the motion planning problem as an optimization problem which can be simplified and efficiently solved by numerical algorithm. Using SDF as environment representation and proposing a multi-terminal path searching method, we achieved more efficient trajectory generation for tractor-trailer robots in unstructured environments. 


In the future, we will further improve the adaptability of our algorithms to cope with more dynamic environments. In addition, integrating technologies from machine learning or graphics to achieve smarter autonomous navigation, such as allowing tractor-trailer robots to acquire semantic information using GRB cameras to have more interaction with real-world environments, or minimizing sweep volume of the trajectory for reducing collision risk\cite{xlh_swept} are also promising directions to explore. Through these efforts, we believe that the application prospects for tractor-trailer robots in various transportation and logistics tasks will become even more expansive.

\appendix
\subsection{Analysis on Optimization}
\label{appendix}
In our pipeline, the PHR-ALM\cite{alm} is adopted as the numerical solver of optimization problems. It solves the original problem as follows:
\begin{align}
\min_x&\quad f(x)\label{prob:alm_raw}\\
\text{s.t.}&\quad h_i(x)=0\quad\quad i=1,2,...,m\nonumber
\end{align}
where $h(x)=[h_1(x),h_2(x),...,h_m(x)]^\text{T}$. The detailed steps of the PHR-ALM are shown in Algorithm~\ref{alg:phr_alm}, where the function FindSolution($\mathcal{L},x_0$) represents solving the minimization problem iteratively from starting point $x_0$ with the objective function $\mathcal{L}$. In work ~\cite{optimization}, a key property about PHR-ALM is mentioned, as shown in Lemma~\ref{lemma}. Through this lemma, we are able to prove the sufficient condition regarding the convergence of PHR-ALM, as shown in Theorem~\ref{theorem:conv}.

\begin{algorithm}
    \caption{PHR-ALM}
    \label{alg:phr_alm}
    \KwIn{$f,h$ in problem (\ref{prob:alm_raw}); starting point $x_0,\lambda_0$; $\gamma>1,\rho_0>0$.}
    \KwOut{the approximate solution $\hat{x}^*$.}
    \Begin
    {
        \For{$k=0,1,2,...$}
        {
            $\mathcal{L}_A^k\leftarrow f(x)+\lambda_{k}^\text{T}h(x)+\frac{1}{2}\rho_{k}\sum_{i=1}^mh_i^2(x)$\;
            $x_{k+1}=\textnormal{FindSolution}(\mathcal{L}_A^k,x_k)$\;
            \If{\textnormal{Convergence}($f,h,x_{k+1}$)}
            {
                    \textbf{return} $x_{k+1}$.
            }
            $\lambda_{k+1}=\lambda_k+\rho_kh_k(x_{k+1})$\;
            $\rho_{k+1}=\gamma\rho_k$\;
        }
    }
\end{algorithm}

\begin{table*}[t]
	\small
	\centering
	\renewcommand\arraystretch{1.2}
	\caption{\label{tab:kkt} Statistics on the degree of constraints violation.}
        \begin{tabular}{c|llcllllll}
\hline
N &
  \multicolumn{1}{c}{$\lVert K_{err}\rVert_\infty$} &
  \multicolumn{1}{c}{Min. $\dot{\mathfrak{s}}$} &
  PCT. $\delta_s$ &
  \multicolumn{1}{c}{PCT. $v_0$} &
  \multicolumn{1}{c}{PCT. $a$} &
  \multicolumn{1}{c}{PCT. $a_\text{lat}$} &
  \multicolumn{1}{c}{PCT. $\kappa$} &
  \multicolumn{1}{c}{PCT. SDF} &
  \multicolumn{1}{c}{$\max(d_e)$} \\ \hline
1 & 0.100 & 0.0423 & 0\% & 1.24\% & 1.13\% & 1.24\% & 1.74\% & 1.99\% & 3.60cm \\
2 & 0.100 & 0.0471 & 0\% & 1.24\% & 1.07\% & 1.20\% & 1.74\% & 1.99\% & 2.83cm \\
3 & 0.100 & 0.0140 & 0\% & 1.23\% & 1.02\% & 1.23\% & 1.75\% & 1.84\% & 3.00cm \\ \hline
\end{tabular}
\end{table*}

\begin{lemma}
Let $x^*$ be a local solution of problem (\ref{prob:alm_raw}) at which the LICQ is satisfied (that is, the Jacobian $\nabla h(x^*)$ is row full rank), and there is a Lagrange multipiler vector $\lambda^*$, such that $\nabla_x\mathcal{L}(x^*,\lambda^*)=0$. Suppose also that: $\forall w\in\{w|\nabla h_i(x^*)^\text Tw=0,i=1,2,...,m\},w\neq0$, there are $w^\text{T}\nabla_{xx}^2\mathcal{L}(x^*,\lambda^*)w>0$. Then there exist positive scalars $\bar\rho,\delta,\epsilon$, and $M$, such that: \\
\textbf{For} all $\lambda_k$ and $\rho_k$ satisfying $\lVert\lambda_k-\lambda^*\rVert\le\rho_k\delta, \rho_k>\bar\rho$, the problem:
\begin{align}
\min_x&\quad \mathcal{L}_A(x,\lambda_k,\rho_k)=f(x)+\lambda_{k}^\text{T}h(x)+\frac{1}{2}\rho_{k}\sum_{i=1}^mh_i^2(x)\label{prob:alm_inner}\\
\text{s.t.}&\quad \lVert x-x^*\rVert\le\epsilon\nonumber
\end{align}
has a unique solution $\hat x$, and $\lVert \hat x-x^*\rVert\le M\lVert\lambda_{k}-\lambda^*\rVert/\rho_k,\lVert \hat\lambda-\lambda^*\rVert\le M\lVert\lambda_{k}-\lambda^*\rVert/\rho_k$. Where $\mathcal{L}(x,\lambda)=f(x)+\lambda^\text{T}h(x),\hat\lambda=\lambda_k+\rho_kh(\hat x)$\label{lemma}
\end{lemma}

\begin{theorem}
Let $x^*,\lambda^*$ and the positive scalars $\bar\rho,\delta,\epsilon,M$ satisfy the conditions in Lemma~\ref{lemma}. Suppose that we have $\lVert\lambda_0-\lambda^*\rVert\le\rho_0\delta,\rho_0>\bar\rho$. If the minimizer $x_{k+1}$ of $\mathcal{L}_A^k$  starting at $x_k$ in Algorithm~\ref{alg:phr_alm} satisfy that $\lVert x_{k+1}-x^*\rVert\le\epsilon$. We have $\lim_{{k\rightarrow\infty}}\lVert x_{k+1}-x^*\rVert=0$.
\label{theorem:conv}
\end{theorem}
\begin{proof}
Since in Algorithm~\ref{alg:phr_alm}, at each step of solving the subproblem about $\mathcal{L}_A^k$, we are guaranteed that $\lVert x_{k+1}-x^*\rVert\le\epsilon$, $x_{k+1}$ is naturally a solution to problem (\ref{prob:alm_inner}) as well. Then, by Lemma~\ref{lemma}, the solution $\hat{x}$ of problem (\ref{prob:alm_inner}) is unique, so $\hat{x}=x_{k+1}$. Thus we have
\begin{align}
\lim_{{k\rightarrow\infty}}\lVert x_{k+1}-x^*\rVert&\le\lim_{{k\rightarrow\infty}}\frac{M}{\rho_k}\lVert\lambda_{k}-\lambda^*\rVert\nonumber\\
&\le\lim_{{k\rightarrow\infty}}\frac{M^2}{\rho_k\rho_{k-1}}\lVert\lambda_{k-1}-\lambda^*\rVert\nonumber\\
&\le...\le\lim_{{k\rightarrow\infty}}(\prod_{i=0}^{k}\frac{M}{\rho_{i}})\lVert\lambda_{0}-\lambda^*\rVert=0,\nonumber
\end{align}
where the last equality sign holds because by Algorithm~\ref{alg:phr_alm}, $\rho_{k+1}=\gamma\rho_k=...=\gamma^{k+1}\rho_0$. When $k\rightarrow\infty,\rho_k\rightarrow\infty$.
\end{proof}

In our pipeline, we solve the problem of the following form: 
\begin{align}
\min_{x} &\quad f(x) \label{prob:gcopt}\\
\text {s.t.} &\quad h_i(x)=0,i=1,2,...,m, \nonumber\\
&\quad g_j(x)\le0,j=1,2,...,r,\nonumber
\end{align}
where $g(x)=[g_1(x),g_2(x),...,g_r(x)]^\text{T}$ are inequality constraints. In practice, we convert the problem (\ref{prob:gcopt}) into the following equivalent problem to match the solvable form as in problem (\ref{prob:alm_raw}). 
\begin{align}
\min_{x,s} &\quad f(x,s)\\
\text {s.t.} &\quad h_i(x)=0,i=1,2,...,m,\nonumber\\
&\quad g_j(x)+s_j^2=0,j=1,2,...,r.\nonumber
\end{align}
As for the subproblem about $\mathcal{L}_A^k$ in Algorithm~\ref{alg:phr_alm}, we solve it using L-BFGS\cite{lbfgs}, a efficient quasi-Newton method for solving largescale unconstrained optimization problems. To check whether the constraints are sufficiently satisfied, we add conditions related to the infinity norm of the constraints to the convergence test in Algorithm~\ref{alg:phr_alm}, i.e., whether $\lVert h(x_{k+1})\rVert_\infty\le\epsilon_\text{cond}$, where $\epsilon_\text{cond}$ is a positive constant.

Here, we provide some numerical statistics for the proposed algorithm to show the degree to which constraints are satisfied, where the parameters are set the same as in the benchmark comparison experiments. Table~\ref{tab:kkt} illustrates the degree of constraint violation for the solutions that successfully converged in the benchmark comparisons, where $\lVert K_{err}\rVert_\infty$ denotes the infinite norm of the equality constraint function introduced by the kinematic model of the tractor-trailer robot, corresponding to Eq.(\ref{con:eqc}). Min. $\dot{\mathfrak{s}}$ denotes the smallest $\dot{\mathfrak{s}}$, corresponding the constraint $\dot{\mathfrak{s}}\ge0$. PCT. $x$ denote the percentage of constraint violations related to $x$. 
\begin{align}
\text{PCT}. \delta_s&=\frac{\max(\delta_+-\min(\boldsymbol{\delta}_s),0) }{\delta_+}\times100\%,\\
\text{PCT}. v_0&=\frac{\max(\lVert\boldsymbol{v}_0\rVert_\infty-v_\text{mlon},0)}{v_\text{mlon}}\times100\%,
\end{align}
\begin{align}
\text{PCT}. a&=\frac{\max(\lVert\boldsymbol{a}\rVert_\infty-a_\text{mlon},0)}{a_\text{mlon}}\times100\%,\\
\text{PCT}. a_\text{lat}&=\frac{\max(\lVert\boldsymbol{a}_\text{lat}\rVert_\infty-a_\text{mlat},0)}{a_\text{mlat}}\times100\%,\\
\text{PCT}. \kappa&=\frac{\max(\lVert\boldsymbol{\kappa}\rVert_\infty-\kappa_\text{max},0)}{\kappa_\text{max}}\times100\%,\\
\text{PCT. SDF}&=\frac{\max(r_s-\text{min}(\boldsymbol{d}),0)}{r_\text{s}}\times100\%.
\end{align} 
$\boldsymbol{\delta}_s$ is the vector consisting of the value $\dot x_0^2(\mathfrak{s})+\dot y_0^2(\mathfrak{s})$ at the discrete trajectory points, corresponding to Condition (39). $\boldsymbol{v}_0,\boldsymbol{a},\boldsymbol{a}_\text{lat},\boldsymbol{\kappa}$ denote vectors consisting of longitudinal velocities, longitude accelerations, latitude accelerations, and curvatures of the trajectory points, respectively. $\boldsymbol{d}$ is the vector consisting of the SDF values for each vehicle center of the robot at the discrete trajectory points. $d_e$ is the maximum distance of the robot end state from the target area. $\delta_+,v_\text{mlon},a_\text{mlon},a_\text{mlat},\kappa_\text{max},r_s$ are constants corresponding to different limitations. 

\newlength{\bibitemsep}\setlength{\bibitemsep}{0.00\baselineskip}
\newlength{\bibparskip}\setlength{\bibparskip}{0pt}
\let\oldthebibliography\thebibliography
\renewcommand\thebibliography[1]{
    \oldthebibliography{#1}
    \setlength{\parskip}{\bibitemsep}
    \setlength{\itemsep}{\bibparskip}
}
\bibliography{references}

\begin{IEEEbiography}[{\includegraphics[width=1in,height=1.15in,clip,keepaspectratio]{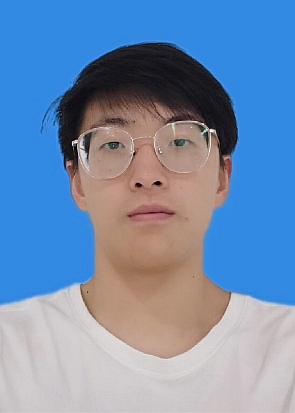}}]
    {Long Xu}
    received the B.Eng. degree in automation from Zhejiang University, Hangzhou, China, in 2022, where he is currently pursuing the Ph.D. degree in electronic information. His research interests include motion planning and robot learning.
\end{IEEEbiography}
\vspace{-1.5cm} 
\begin{IEEEbiography}[{\includegraphics[width=1in,height=1.15in,clip,keepaspectratio]{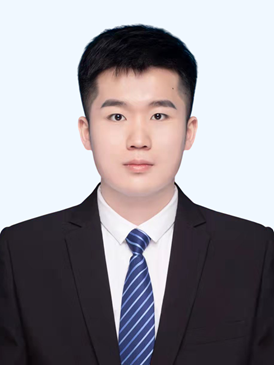}}]
    {Kaixin Chai}
    graduated from Xi'an Jiaotong University in 2022, with a major in Energy and Power Engineering. After graduation, he began his research in robotics and worked as a visiting student at Zhejiang University, City University of Hong Kong, and Korea Advanced Institute of Science \& Technology. His research interests include perception-aware planning, mobile manipulation, and humanoid whole-body control.
\end{IEEEbiography}
\vspace{-1.5cm} 
\begin{IEEEbiography}[{\includegraphics[width=1in,height=1.15in,clip,keepaspectratio]{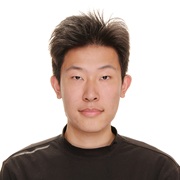}}]
    {Boyuan An}
    is currently an undergraduate student in Robotics Engineering at Zhejiang University, expected to graduate in 2027. He is presently working as a research intern at the AI4CE Lab, New York University. His research interests include reinforcement learning, computer vision, and autonomous robotic systems.
\end{IEEEbiography}
\vspace{-1.5cm}
\begin{IEEEbiography}[{\includegraphics[width=1in,height=1.15in,clip,keepaspectratio]{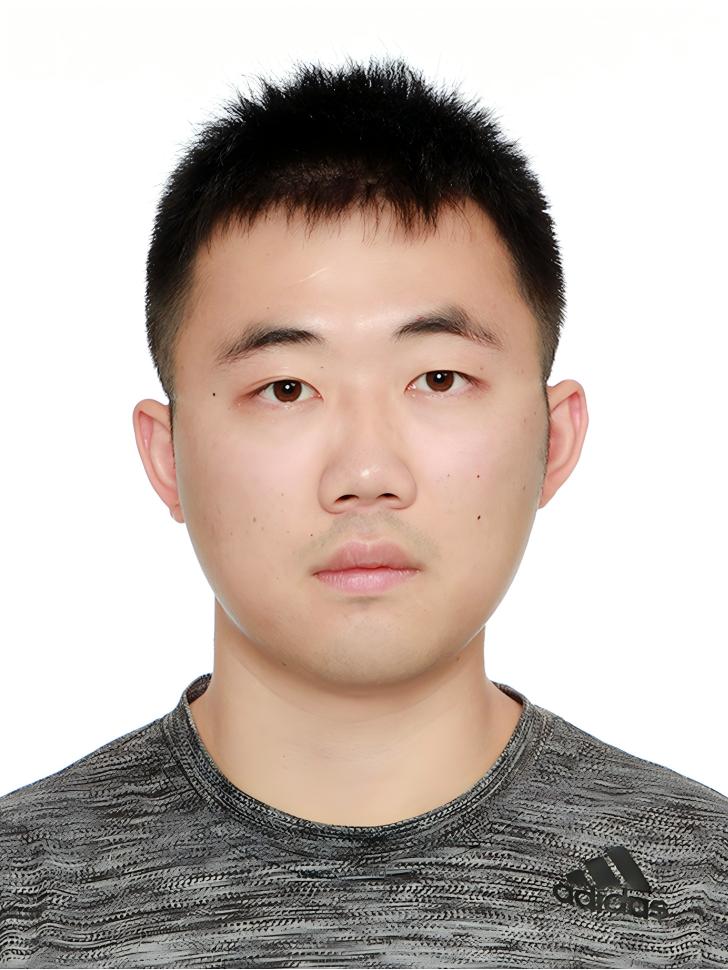}}]
    {Shuhang Ji}
    received the B.Eng. degree in control science and engineering from Zhejiang University, Hangzhou, China, in 2025, currently working as a research assistant. His research interests include aerial manipulator, robot system and control.
\end{IEEEbiography}
\vspace{-1.5cm}
\begin{IEEEbiography}[{\includegraphics[width=1in,height=1.15in,clip,keepaspectratio]{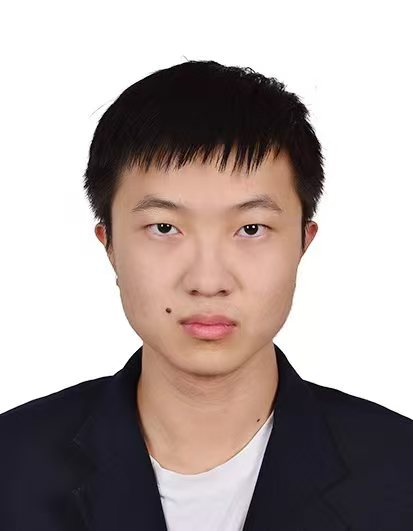}}]
    {Zhenyu Hou}
    received the B.Eng. degree in Cyber Security from the Civil Aviation University of China, Tianjin, China, in 2024. He is currently a Research Assistant with the Huzhou Institute, Zhejiang University. His research interests include navigation and planning for mobile robots.
\end{IEEEbiography}
\vspace{-1.5cm}
\begin{IEEEbiography}[{\includegraphics[width=1in,height=1.15in,clip,keepaspectratio]{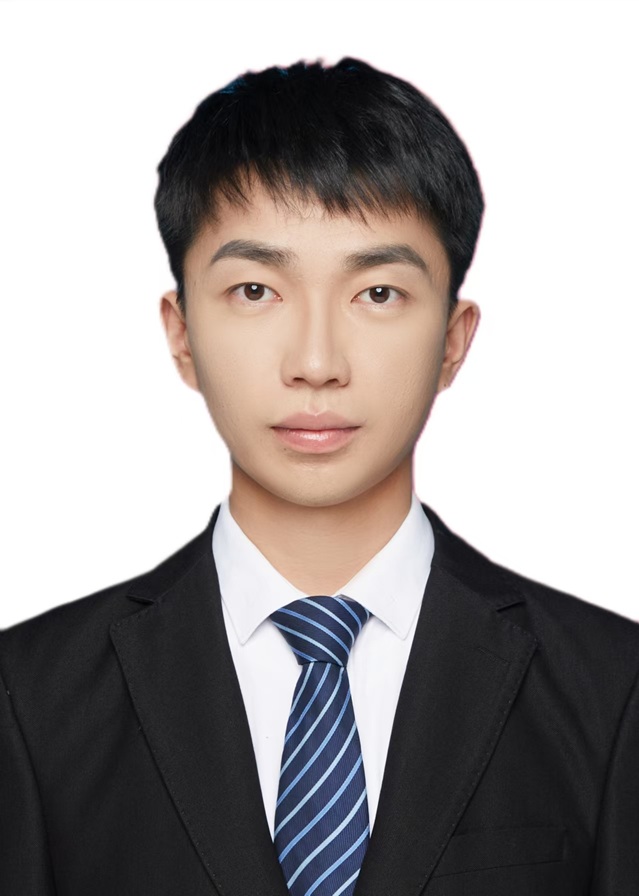}}]
    {Jiaxiang Gan}
    received the M.Eng. degree in Control Engineering from North China Electric Power University, Baoding, China, in 2025. His research interests include Path Planning and UAV control.
\end{IEEEbiography}
\vspace{-1.5cm}
\begin{IEEEbiography}[{\includegraphics[width=1in,height=1.15in,clip,keepaspectratio]{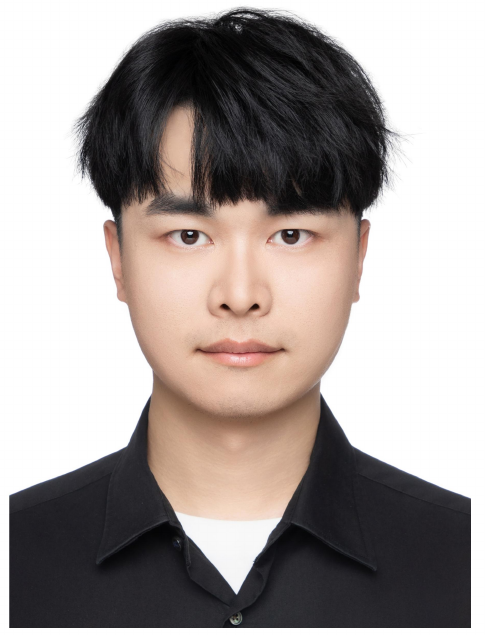}}]
    {Yuan Zhou}
    received the B.Eng. degree in mechanical engineering in 2022 from Northeastern University, Shenyang, China. He is currently working toward the Ph.D. degree in control science and engineering at Zhejiang University, Hangzhou, China. His research interests include formation planning for multirobot systems, swarm intelligence.
\end{IEEEbiography}
\vspace{-1.5cm}
\begin{IEEEbiography}[{\includegraphics[width=1in,height=1.15in,clip,keepaspectratio]{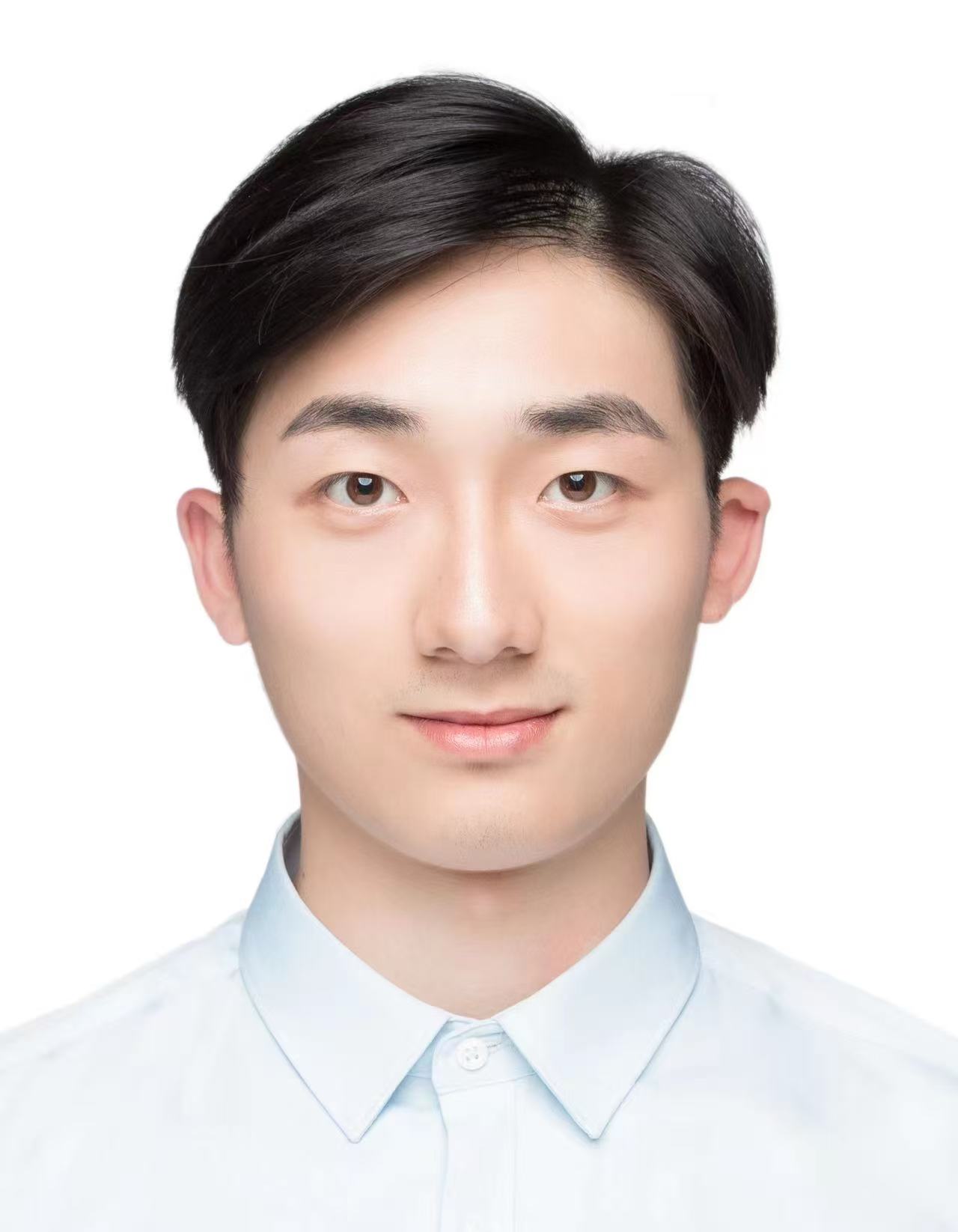}}]
    {Qianhao Wang}
    received the M.Eng. degree in control science and engineering in 2022 from Zhejiang University, Hangzhou, China, where he is currently working toward the Ph.D. degree in control science and engineering. His research interests include motion planning, computational geometry, LiDAR SLAM, and autonomous navigation for aerial robotics.
\end{IEEEbiography}
\vspace{-1.5cm}
\begin{IEEEbiography}[{\includegraphics[width=1in,height=1.15in,clip,keepaspectratio]{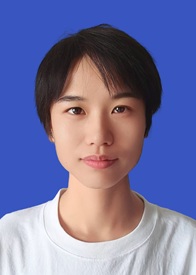}}]
    {Xiaoying Li}
    received the M.Eng. degree in Control Science and Engineering from Harbin Engineering University, China, in 2018. She has worked in both industry and research labs, focusing on robotics applications such as obstacle avoidance, autonomous driving, and robotic manipulation. Her research interests include motion planning and trajectory optimization for autonomous systems, robot navigation on uneven terrain, and intelligent control.
\end{IEEEbiography}
\vspace{-1.5cm}
\begin{IEEEbiography}[{\includegraphics[width=1in,height=1.15in,clip,keepaspectratio]{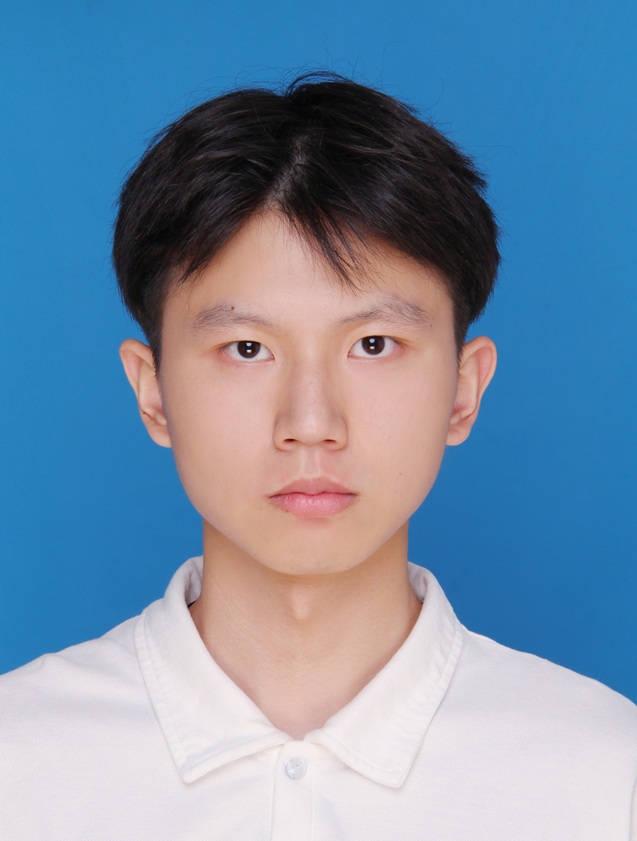}}]
    {Junxiao Lin}
    received the B.Eng. degree in mechanical engineering from Zhejiang University, Hangzhou, China, in 2023, where he is currently pursuing the M.Phil. degree in control engineering. His research interests include mobile robots, design and control.
\end{IEEEbiography}
\vspace{-1.5cm}
\begin{IEEEbiography}[{\includegraphics[width=1in,height=1.15in,clip,keepaspectratio]{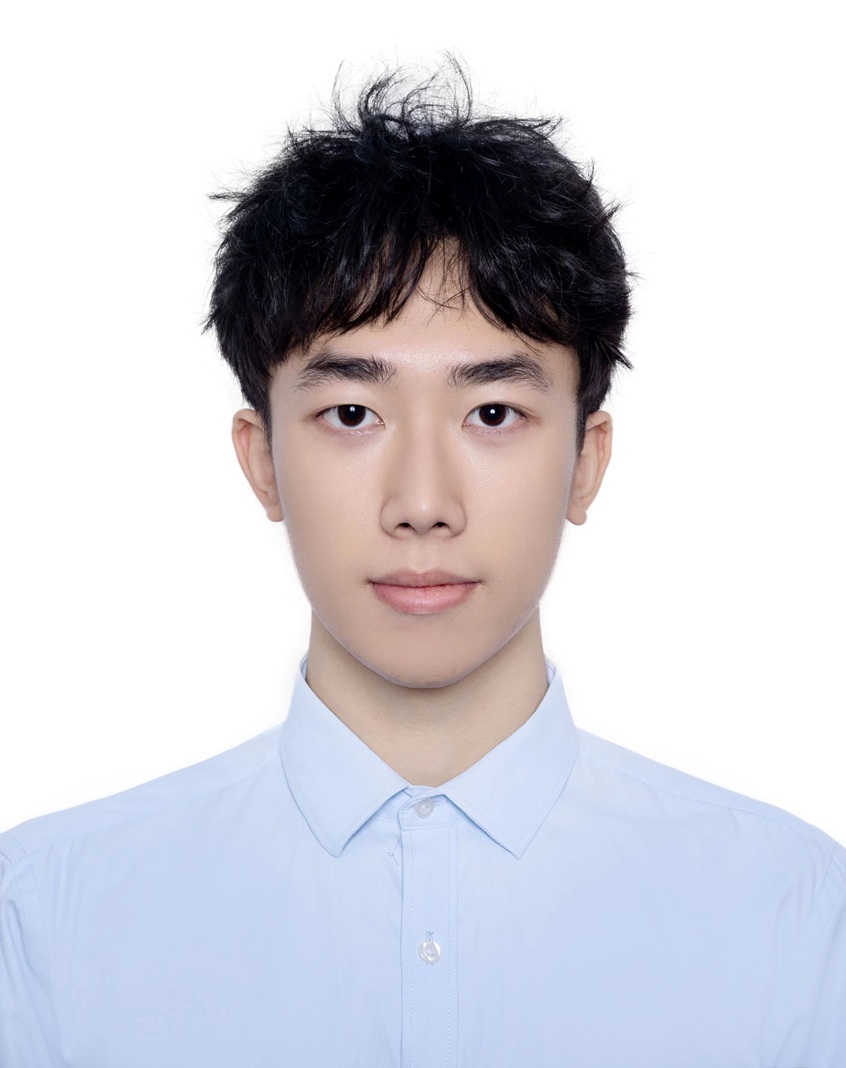}}]
    {Zhichao Han}
    received the B.Eng. degree in Automation from Zhejiang University, Hangzhou, China, in 2021. Currently, he is actively pursuing a Ph.D. degree in the Fast Lab at Zhejiang University, where he is under the supervision of Prof. Fei Gao. His research interests include motion planning and robot learning.
\end{IEEEbiography}
\vspace{-1.5cm}
\begin{IEEEbiography}[{\includegraphics[width=1in,height=1.15in,clip,keepaspectratio]{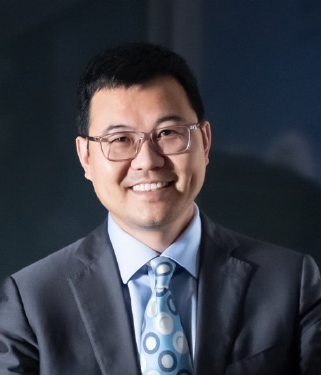}}]
    {Chao Xu}
    received the Ph.D. degree in mechanical engineering from Lehigh University in 2010. He is currently the Associate Dean and a Professor with the College of Control Science and Engineering, Zhejiang University. He is the Inaugural Dean of ZJU Huzhou Institute. His research expertise is flying robotics and control-theoretic learning. He has published over 100 articles in international journals, including Science Robotics and Nature Machine Intelligence. He will join the organization committee of the IROS-2025 in Hangzhou.
\end{IEEEbiography}
\vspace{-1.5cm}
\begin{IEEEbiography}[{\includegraphics[width=1in,height=1.15in,clip,keepaspectratio]{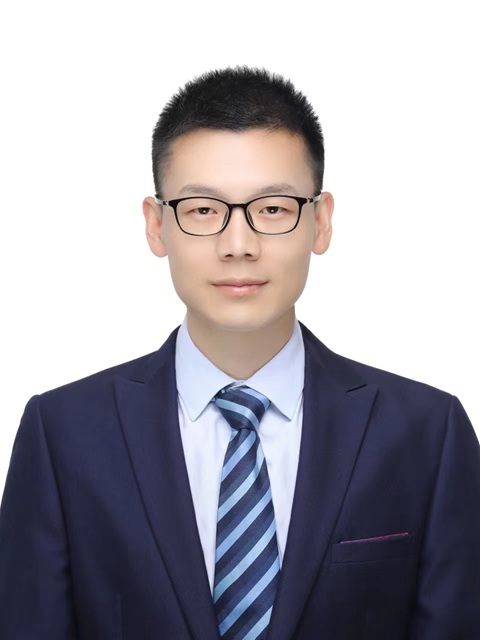}}]
    {Yanjun Cao}
    received his Ph.D. degree in computer and software engineering from the University of Montreal, Polytechnique Montreal, Canada, in 2020. He is currently an associate researcher at the Huzhou Institute of Zhejiang University, as a PI in the Center of Swarm Navigation. He leads the Field Intelligent Robotics Engineering group of the Field Autonomous System and Computing Lab. His research focuses on key challenges in multi-robot systems, such as collaborative localization, autonomous navigation, perception and communication.
\end{IEEEbiography}
\vspace{-1.5cm}
\begin{IEEEbiography}[{\includegraphics[width=1in,height=1.15in,clip,keepaspectratio]{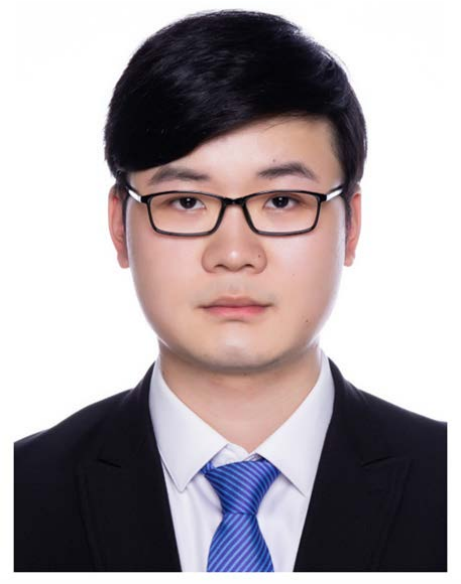}}]
    {Fei Gao}
    received the Ph.D. degree in electronic and computer engineering from the Hong Kong University of Science and Technology, Hong Kong, in 2019. He is currently a tenured associate professor at the Department of Control Science and Engineering, Zhejiang University, where he leads the Flying Autonomous Robotics (FAR) group affiliated with the Field Autonomous System and Computing (FAST) Laboratory. His research interests include aerial robots, autonomous navigation, motion planning, optimization, and localization and mapping.
\end{IEEEbiography}

\end{document}